\documentclass[final]{cvpr}

\usepackage{times}
\usepackage{graphicx}
\usepackage{epsfig}
\usepackage{amssymb}
\usepackage{amsmath}
\usepackage{subfigure}
\usepackage{bbding}
\usepackage{booktabs}
\usepackage{lipsum}

\newcommand\blfootnote[1]{%
  \begingroup
  \renewcommand\thefootnote{}\footnote{#1}%
  \addtocounter{footnote}{-1}%
  \endgroup
}

\usepackage{paralist}

\usepackage[pagebackref=true,breaklinks=true,colorlinks,bookmarks=false]{hyperref}
\usepackage[ruled]{algorithm2e}

\def\cvprPaperID{2115} % *** Enter the CVPR Paper ID here
\def\confYear{CVPR 2021}

\begin{document}

%%%%%%%%% TITLE
\title{Deep Graph Matching under Quadratic Constraint}

\author{Quankai Gao$^1$, Fudong Wang$^1$, Nan Xue$^1$, Jin-Gang Yu$^{2}$, Gui-Song Xia$^{1}$\\
$^1$Wuhan University, Wuhan, China.\\
$^2$South China University of Technology, Guangzhou, China.\\
{\tt\small \{quankaigao, fudong-wang, xuenan, guisong.xia\}@whu.edu.cn},
{\tt\small jingangyu@scut.edu.cn}
% For a paper whose authors are all a  t the same institution,
% \blfootnote{test}

%\and
}

\maketitle

\blfootnote{The study of this paper is funded by the National Natural Science Foundation of China (NSFC) under grant contracts No.61922065, No.61771350 and No.41820104006 and 61871299. It is also supported by Supercomputing Center of Wuhan University.}
\blfootnote{
Corresponding author: Gui-Song Xia (guisong.xia@whu.edu.cn).}

%%%%%%%%% ABSTRACT
\begin{abstract}
    Recently, deep learning based methods have demonstrated promising results on the graph matching problem, by relying on the descriptive capability of deep features extracted on graph nodes. However, one main limitation with existing deep graph matching (DGM) methods lies in their ignorance of explicit constraint of graph structures, which may lead the model to be trapped into local minimum in training. In this paper, we propose to explicitly formulate pairwise graph structures as a \textbf{quadratic constraint} incorporated into the DGM framework. The quadratic constraint minimizes the pairwise structural discrepancy between graphs, which can reduce the ambiguities brought by only using the extracted CNN features.
    Moreover, we present a differentiable implementation to the quadratic constrained-optimization such that it is compatible with the unconstrained deep learning optimizer. To give more precise and proper supervision, a well-designed false matching loss against class imbalance is proposed, which can better penalize the false negatives and false positives with less overfitting. Exhaustive experiments demonstrate that our method achieves competitive performance on real-world datasets.
\end{abstract}

%%%%%%%%% BODY TEXT
\section{Introduction}
Graph matching aims to find an optimal one-to-one node correspondence between graph-structured data, which has been widely used in many tasks~\cite{berg2005shape,brendel2011learning,conte2004thirty,gasse2019exact,jiang2010linear,szeliski2010computer}. By integrating the similarity between nodes and edges in a combinatorial fashion, graph matching is often mathematically formulated as a quadratic assignment problem (QAP)~\cite{loiola2007survey}. QAP is known to be NP-hard~\cite{hartmanis1982computers}, and various approximation techniques~\cite{leordeanu2005spectral,leordeanu2009integer,liu2014graph,schellewald2001evaluation} have been proposed to make it computationally tractable.

\begin{figure}[t!]
\centering
\subfigure[DGM {\em without} quadratic constraint]{
\includegraphics[height=0.25\linewidth, width=0.7\linewidth]{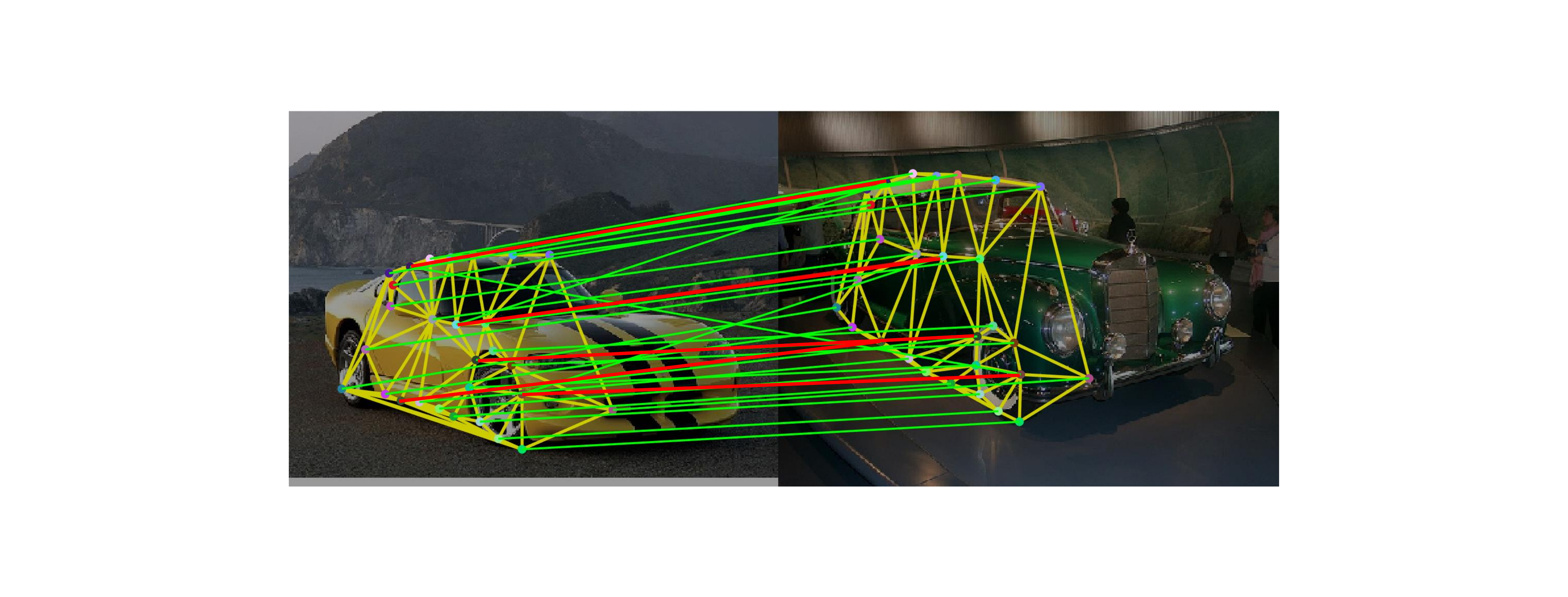}
\includegraphics[height=0.25\linewidth]{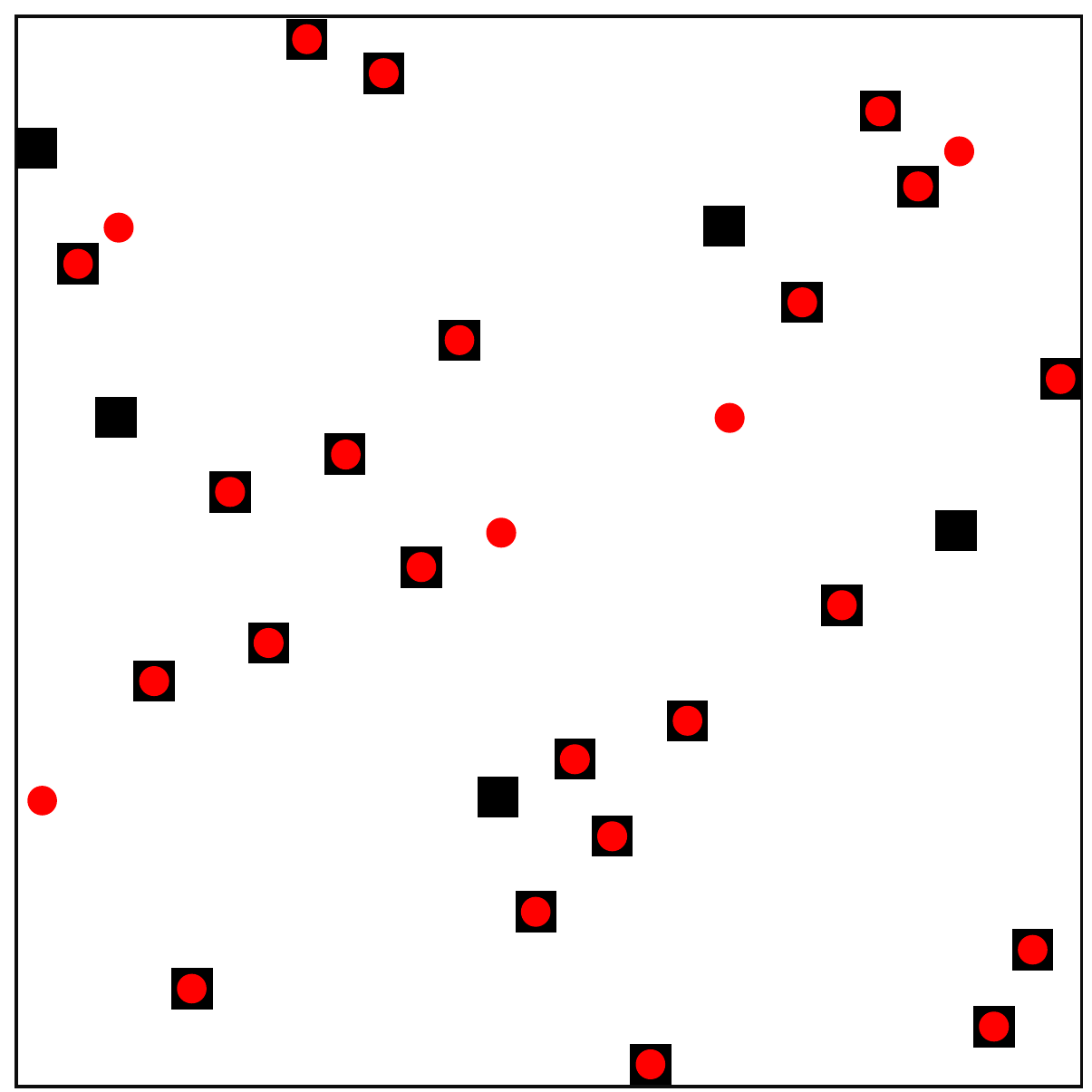}
}\\
\vspace{-3mm}
\subfigure[DGM {\em with} quadratic constraint (our method)]{
\includegraphics[height=0.25\linewidth, width=0.7\linewidth]{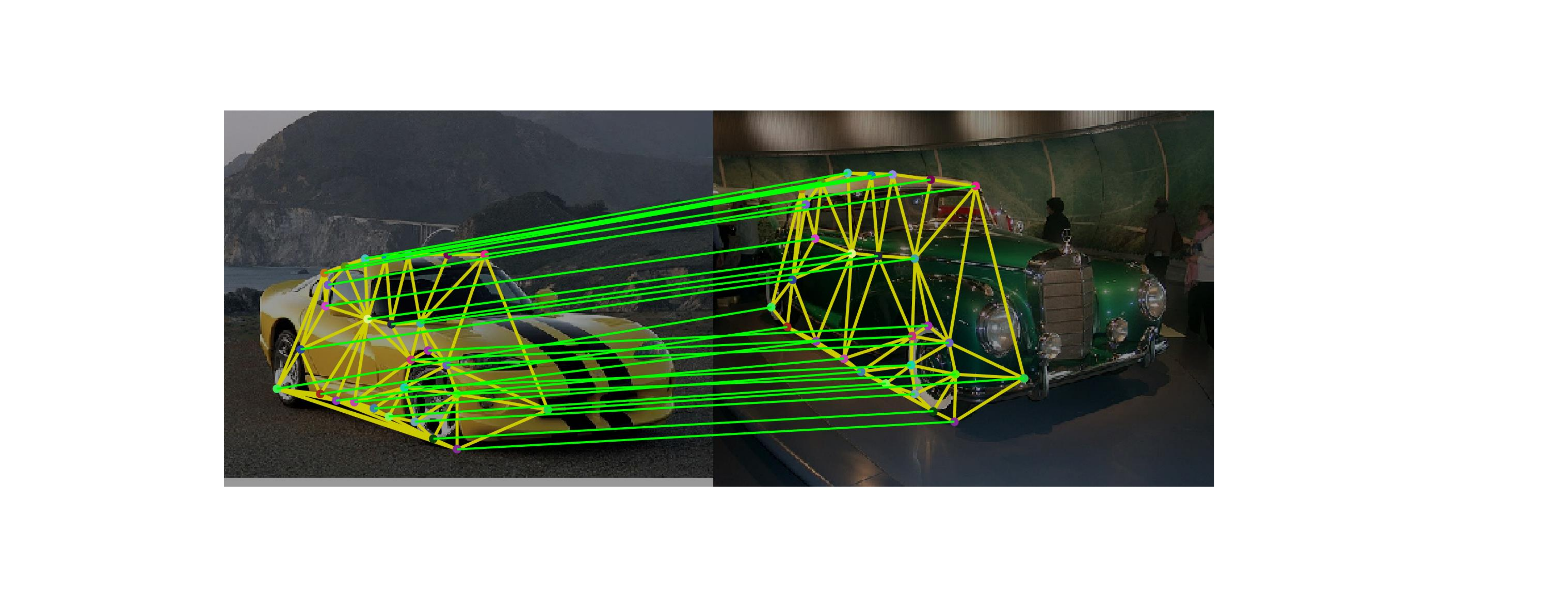}
\includegraphics[height=0.25\linewidth]{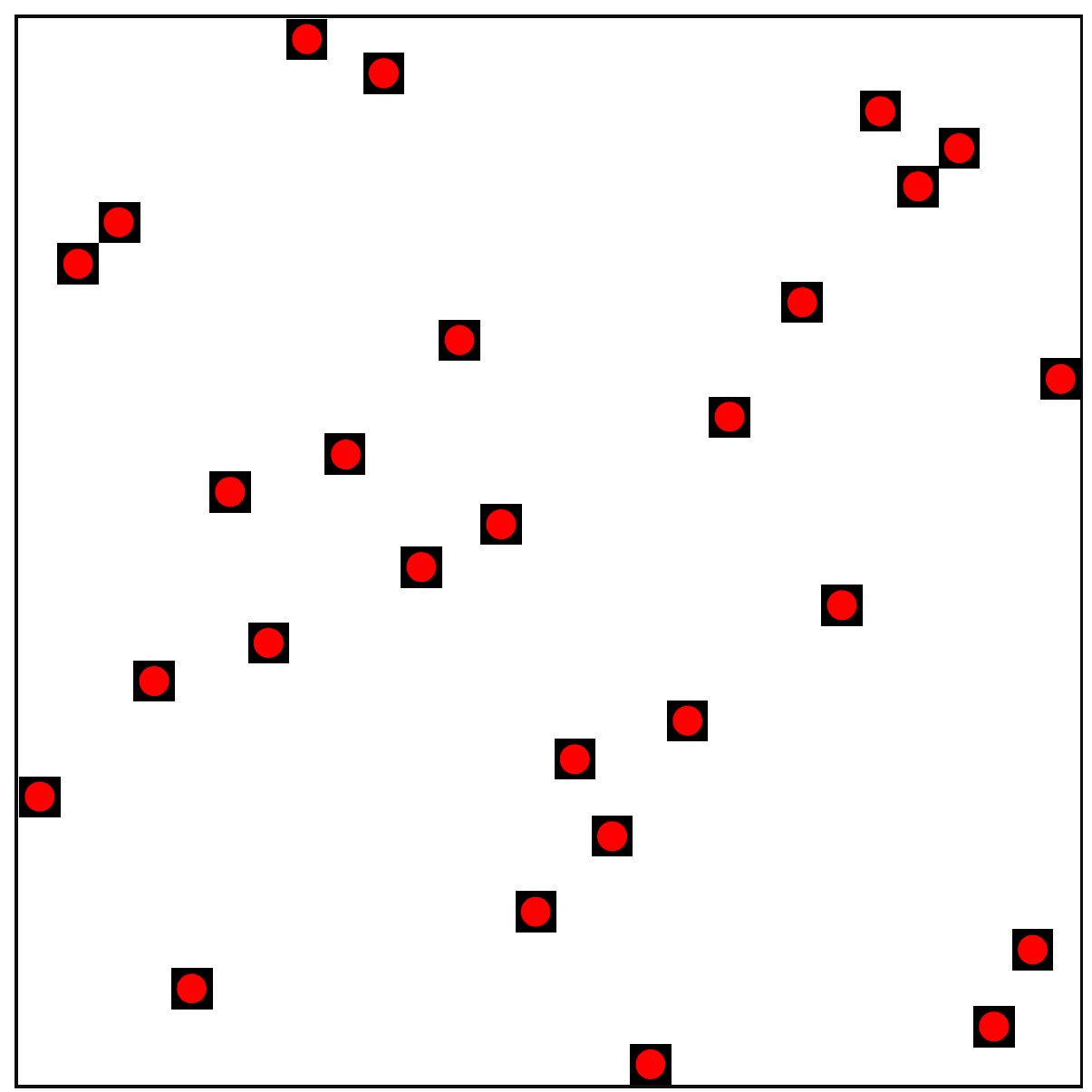}
}
\caption{Deep graph matching with/without quadratic constraint. Matching results are in left and the predicted (black) {\em v.s.} the ground truth (red) correspondence matrices are in right.}
\vspace{-3mm}
\label{fig:QC}
\end{figure}

\begin{figure*}[t]
\centering
   \includegraphics[width=0.87\linewidth]{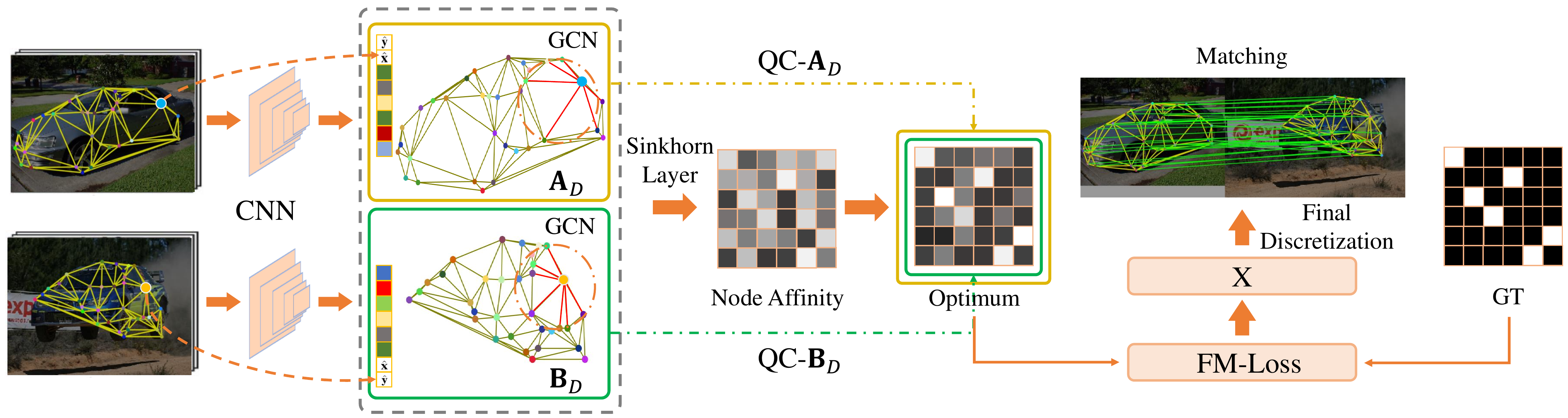}
%\vspace{-1mm}
   \caption{Overview of our proposed architecture for deep graph matching under quadratic constraint. Node attributes consisting geometric prior and deep features are refined to build the initial node affinity matrix, which is followed by a Sinkhorn layer and then further optimized under quadratic constraint (QC). Loss between the prediction and the ground truth (GT) is calculated by the proposed false matching loss (FM-Loss).}
   \vspace{-3mm}
\label{fig:overview}
\end{figure*}

Until recently, deep graph matching (DGM) methods give birth to many more flexible formulations~\cite{fey2020deep,rolinek2020deep,wang2019learning,yu2020learning} besides traditional QAP. DGM aims to learn the meaningful node affinity by using deep features extracted from convolutional neural network. To this end, many existing DGM methods~\cite{rolinek2020deep,wang2019learning,yu2020learning} primarily focus on the feature modeling and refinement for more accurate affinity construction. The feature refinement step is expected to capture the implicit structure information~\cite{wang2019learning} encoded in learnable parameters.
%Since the extracted raw features encoding semantic information of background image are independent of graph structures, the feature refinement step is expected to capture the implicit structure information~\cite{wang2019learning} encoded in learnable parameters.
However, nodes with similar deep features are hard to distinguish from each other in deep graph matching, while their structure contexts may be very different. Moreover, the implicit structure information is not specific enough, which is insufficient to clearly represent the structural discrepancy over graphs ({\em e.g.}, Fig.~\ref{fig:QC}(a)).

In traditional graph matching, it is common to
incorporate pairwise structures into the formulation to enhance matching accuracy~\cite{leordeanu2005spectral}, which inspired us to consider quadratic structural constraint in deep graph matching to maximize the adjacency consensus and achieve global consistency. More precisely, we use the pairwise term of Koopmans-Beckmann’s QAP~\cite{loiola2007survey} as our quadratic constraint to minimize the adjacency discrepancy of graphs to be matched ({\em e.g.}, Fig.~\ref{fig:QC}(b)). To this end, we present a modified Frank-Wolfe algorithm~\cite{lacoste2015global}, which is a differentiable optimization scheme {\em w.r.t.} learnable parameters in our model and the relaxed Koopmans-Beckmann’s QAP.

Another important issue of deep graph matching is class imbalance. Concretely, the result of a graph matching task is usually represented as a permutation matrix, where only a small portion of the entries take the value of one representing the pairs to be matched while the rest are zero-valued, leading to the imbalance between matched and unmatched entries. In case of such class imbalance, it will be problematic to establish the loss function between predicted matching matrices and ground truth matrices by using the conventional cross-entropy-type loss functions (see Section~\ref{subsec:fmloss} for details). To our best knowledge, there is no loss function specifically designed for deep graph matching to take care of the class imbalance issue so far. To this end, we design a novel loss function for deep graph matching, called False Matching Loss, which will be experimentally shown to be better for dealing with class imbalance and overfitting in compared with previous works.

Our main contributions are highlighted as follows:
\begin{itemize}
    \item[-] We explicitly introduce quadratic constraint with our constructed geometric structure into deep graph matching, which can further revise wrong matches by minimizing structure discrepancy over graphs.
    \item[-] We present a differentiable optimization scheme in training to approach the objective such that it is compatible with the unconstrained deep learning optimizer.
    \item[-] We propose a novel loss function focusing on false matchings, {\em i.e.} false negatives and false positives, to better lead the parameter update against class imbalance and overfitting.
\end{itemize}

\section{Preliminaries and Related Work}
For better understanding, this section will revisit some preliminaries and related works on both traditional combinatorial graph matching and deep graph matching.

\subsection{Combinatorial graph matching}
Graph matching aims to build the node-to-node correspondence between the given two graphs $\mathcal{G}_A = \{V_A, E_A\}$ with $|V_A|=n$ and $\mathcal{G}_B = \{V_B, E_B\}$ with $|V_B|=m$, where $V$ we denote as the set of nodes and $E$ as the set of edges. By denoting $\mathbf{X}$ as the correspondence matrix indicating the matching between two graphs $\mathcal{G}_A$ and $\mathcal{G}_B$, {\em i.e.}, $X_{ij}=1$ means the $i$-th node in $V_A$ matches to the $j$-th node in $V_B$, $X_{ij}=0$ otherwise, one well-known form of graph matching with combinatorial nature can be written as:
\begin{align}
\label{eq:KBQAP}
\min_{\mathbf{X}}\quad ||\mathbf{A}-\mathbf{XBX}^T||^2_F-{\rm tr}(\mathbf{X}^T_u\mathbf{X})\\
\mathbf{X}\in\{0,1\}^{n\times m}, \mathbf{X1}_{n}=\mathbf{1}_{m}, \mathbf{X}^T\mathbf{1}_{n}\leq\mathbf{1}_{n} \nonumber
\end{align}
where $\mathbf{A}\in \mathbb{R}^{n\times n}$, $\mathbf{B}\in \mathbb{R}^{m\times m}$ are adjacency matrices encoding the pairwise information of edges in graphs $\mathcal{G}_A$ and $\mathcal{G}_B$, respectively. $\mathbf{X}_u\in\mathbb{R}^{n\times m}$ measures the node similarities between two graphs. $||\cdot||_F$ is the Frobenius norm. Generally, Eq.~\eqref{eq:KBQAP} can be cast as a quadratic assignment problem called Koopmans-Beckmann's QAP~\cite{loiola2007survey}. One can find more details in \cite{yan2016short}.

In the previous works \cite{gold1996graduated,wfd2020frgm,zhou2015factorized} following this combinatorial graph matching formulation, the node similarity $\mathbf{X}_u$ and adjacency matrices $\mathbf{A},\mathbf{B}$ are usually calculated with some specifically designed handcrafted features like SIFT \cite{lowe2004distinctive}, Shape Context \cite{belongie2002shape}, {\em etc}. And then they will solve the objective functions ({\em e.g.}, Eq.~\eqref{eq:KBQAP}) with different discrete or continuous constrained-optimization algorithms \cite{cho2014finding,wfd2020frgm,egozi2012probabilistic,yan2015discrete}. Until recently, the deep learning based graph matching framework has been developed
consisting of learned features and unconstrained optimizer, which will be detailed in next section.

\subsection{Deep graph matching}
$\textbf{Feature extraction.}$ Recently, many works~\cite{fey2020deep,wang2019learning,wang2020combinatorial,yu2020learning,rolinek2020deep} on deep graph matching have been proposed to take advantages of the descriptive capability of high-dimensional deep features as node or edge attributes, which can collect visual information of background images. As a basic setting of these works, the output of CNN layers ${\rm relu4\_2}$ and ${\rm relu5\_1}$ are commonly used, denoted as $\mathbf{U}\in\mathbb{R}^{n\times d}$ and $\mathbf{F}\in\mathbb{R}^{n\times d}$,
\begin{align}
\label{eq:ftU}
\textbf{U} &={\rm align}({\rm CNN_{relu4\_2}}(I), V) \\
\label{eq:ftF}
\textbf{F} &={\rm align}({\rm CNN_{relu5\_1}}(I), V)
\end{align}
where $I$ denotes the input image and $V$ denotes the annotated keypoints. CNN here is a widely used architecture VGG16~\cite{simonyan2014very} initially pretrained on the ImageNet~\cite{russakovsky2015imagenet}. "align" in Eq.~\eqref{eq:ftU} and Eq.~\eqref{eq:ftF} is bi-linear interpolation to approximately assign the output features of a convolution layer to the annotated keypoints on the input image pairs.

$\textbf{Feature modeling and refinement.}$
Since the extracted raw features are independent with the graph structure, various refinement strategies on the raw features are adopted in deep graph matching trying to implicitly utilize the information of graph structure.
As a typical example of non-Euclidean data, a graph with its node and edge attributes can be processed by graph convolutional network (GCN)~\cite{kipf2016semi} under the message passing scheme to update its attributes. Each node attribute is updated by aggregating its adjacency node attributes so that GCN is expected to implicitly capture the structure contexts of each node to some extent. There are also some works~\cite{rolinek2020deep,yu2020learning} using unary node features to model pairwise edge features for matching.

$\textbf{Differentiable optimization.}$
Deep graph matching asks for the model fully-differentiable so that many combinatorial solvers and methods ({\em e.g.}, Hungarian algorithm~\cite{kuhn1955hungarian} and IPFP~\cite{leordeanu2009integer}) are not recommended. Thus, various relaxation approaches become popular, which refer to the recent progress~\cite{rolinek2020deep,wang2019learning,wang2020learning}.
%Some approaches adopt Sinkhorn layer~\cite{adams2011ranking} as the linear assignment solver applying normalization on their soft correspondence matrix iteratively to get a doubly-stochastic version.

Under these settings, deep graph matching can be reformulated as maximizing the node affinity based on the extracted features.

$\textbf{Loss function for deep graph matching.}$ Though many works about deep graph matching have been proposed, there are few thorough discussions about loss functions. The prototype of cross entropy loss has been widely used in deep graph matching, {\em e.g.} permutation loss~\cite{wang2019learning} and its improved version~\cite{yu2020learning}. Instead of directly calculating the linear assignment cost, GMN~\cite{zanfir2018deep} uses ``displacement loss" measuring pixel-wise offset on the image but is shown to have a weaker supervision than cross entropy loss~\cite{wang2019learning}. However, none of the existing works consider the class imbalance that naturally exists in deep graph matching. Besides, overfitting and gradient explosion are always conspicuous on models trained with cross-entropy-type loss functions. For the above reasons, we propose a novel loss function specifically designed for deep graph matching, which not only considers numerical issue but also shows promising performance against overfitting and class imbalance.

\section{DGM under Quadratic Constraint}
We briefly demonstrate our method overview here. As shown in Fig.~\ref{fig:overview}, given the input two images with detected or annotated keypoints as graph nodes, we firstly adopt the CNN features and the coordinates of keypoints to calculate both the initial node attributes and the pairwise structural context as weighted adjacency matrices. By this end, we establish graphs with structural attributes, based on which we explicitly use the weighted adjacency matrices to construct the quadratic constraint, and design a differentiable constrained-optimization algorithm to achieve compatibility with the deep learning optimizer. Since the quadratic constrained-optimization is non-convex and needs a proper initialization, we update the node attributes with weighted adjacency matrices through a GCN module and a differentiable Sinkhorn layer~\cite{adams2011ranking,sinkhorn1967concerning} respectively, to obtain a node affinity matrix between two graphs as the initialization. Finally, the solved optimum will be compared with the ground truth by the proposed false matching loss, which addresses the issue of class imbalance to achieve better performance.

\subsection{Geometric structure for DGM}

{\bf{Unary geometric prior} }
Since our work doesn't focus on deep feature extraction, we follow the previous deep graph matching works to use CNN features described in Eq.~\eqref{eq:ftU} and Eq.~\eqref{eq:ftF}. Moreover, features from different layers of CNN are expected to incorporate both the local and global semantic information of images, we concatenate $\mathbf{U}$ and $\mathbf{F}$ together as $\mathbf{P}_r={\rm cat}(\mathbf{U};\mathbf{F})$ to be the initial node attributes of two graphs.

Since the extracted raw features associated to nodes only have visual information of local patches, to make raw node attributes more discriminative, we add the the normalized 2D Cartesian coordinate $[\hat{\mathbf{x}},\hat{\mathbf{y}}]$ of each node as $\mathbf{P}={\rm cat}(\mathbf{P}_r;[\hat{\mathbf{x}},\hat{\mathbf{y}}])$, which provides a unary geometric prior that can better describe the locations of nodes as a complement to the CNN features.

\begin{figure}[t]
\centering
   \includegraphics[width=0.9\linewidth]{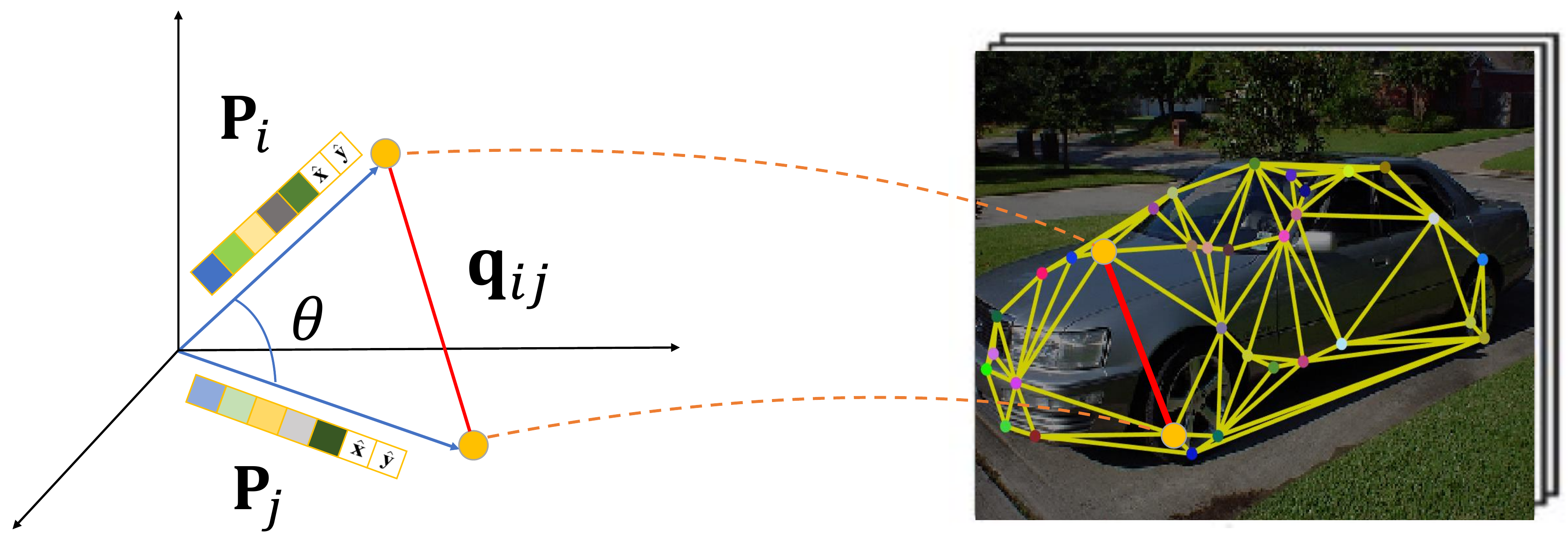}
   \caption{Illustration of the geometric relationship between two nodes and their edge connection in attribute space. Normalized attributes of nodes are represented as coloured squares.}
\label{fig:cosine}
\vspace{-3mm}
\end{figure}

{\bf{Pairwise structural context} }
In deep graph matching, the graph construction is usually based on node coordinates and never consider the visual meaningful features of the background image. For this reason, we introduce deep feature weighted adjacency matrices $\mathbf{A}_D$ and $\mathbf{B}_D$ of the two graphs to learn more proper relations among graph nodes, which are defined as
\begin{equation}
\label{eq:adj}
\mathbf{A}_D=f(\mathbf{P_A})\odot\mathbf{A},
\quad
\mathbf{B}_D=f(\mathbf{P_B})\odot\mathbf{B}
\end{equation}
where $\mathbf{A}\in\mathbb{R}^{n\times n},\mathbf{B}\in\mathbb{R}^{m\times m}$ are the binary adjacency matrices built on coordinates of nodes in two graphs,  $\mathbf{P}_A\in\mathbb{R}^{n\times (d+2)},\mathbf{P}_B\in\mathbb{R}^{m\times (d+2)}$ are the above-mentioned node attributes of two graphs, $\odot$ denotes element-wise product. $f(\mathbf{P})$ can be various commutative function and here we use a linear kernel $f_{i,j}=\mathbf{p}^T_i\mathbf{p}_j$ for simplicity, where $\mathbf{p}_i$ is the $i$-th row of $\mathbf{P}$. As illustrated in Fig.~\ref{fig:cosine}, the geometric meaning of the function $f$ is related to the cosine of the angle between two normalized node attributes, which can be detailed as cos$\theta_{ij}=\frac{\left<\mathbf{p}_i,\mathbf{p}_j\right>}{|\mathbf{p}_i||\mathbf{p}_j|}=\frac{1}{\sqrt{3}}\frac{1}{\sqrt{3}}\left<\mathbf{p}_i,\mathbf{p}_j\right>=\frac{1}{3}\mathbf{p}^T_i\mathbf{p}_j$. By this definition, each element of $\mathbf{A}_D$ and $\mathbf{B}_D$ represents the feature distance between the corresponding nodes while preserving the topology constraints provided by $\mathbf{A}$ and $\mathbf{B}$.

{\textbf{Attributes fusion with GCN} }
There are many convolutional architectures~\cite{fey2018splinecnn,zhou2018graph} for processing irregular data. As a typical example of non-Euclidean data, a graph with its node and edge attributes can be processed by GCN under the message passing scheme to update its attributes. Each node attribute is updated by aggregating its adjacency node attributes so that GCN is expected to capture the structure contexts of each node in an implicit way. With the above interpretation, we adopt GCN that incorporates both update from neighbors and self-update, which can be written as:
\begin{equation}
\label{eq:GCN}
\textbf{P}^{l+1}=\sigma(\textbf{A}_D\textbf{P}^l\textbf{W}^l_r+\textbf{P}^l\textbf{W}^l_s)
\end{equation}
where $\mathbf{W}^l_r,\mathbf{W}^l_s\in\mathbb{R}^{(d+2)\times(d+2)}$ denote learnable parameters of GCN at $l$-th layer. $\sigma$ is the activation function. The updated attributes $\mathbf{P}^{l+1}$ is then used to update $\mathbf{A}_D$ and $\mathbf{B}_D$ by Eq.~\eqref{eq:adj}.

{\bf{Node affinity} }
Since we have the refined attributes of two graphs, the node affinity $\mathbf{K}^l_p\in\mathbb{R}^{n_A\times n_B}$ at $l$-th iteration can be built by a learnable metric:
\begin{equation}
\label{eq:affinity}
\mathbf{K}^l_p={\rm exp}\{{\mathbf{P}^l_A\textbf{W}^l_{{\rm aff}}{\mathbf{P}^l_B}^T}\}
\end{equation}
where $\mathbf{W}^l_{{\rm aff}}$ is a matrix containing learnable parameters. We then adopt Sinkhorn layer taking $\mathbf{K}^l_p$ to the set of doubly-stochastic matrices $\mathcal{D}$, which is the convex hull of the set of permutation matrices $\mathcal{P}$.

\subsection{Quadratic constraint for DGM}
To explicitly utilize the information of graph structures, we formulate the objective as to minimize the pairwise structure discrepancy between the graphs to be matched. As Eq.~\eqref{eq:KBQAP}, Koopmans-Beckmann’s QAP explicitly involves the second-order geometric context in its pairwise term and the optimal solution will minimize the two terms simultaneously. The learnable metric $\mathbf{K}^l_p$ we obtained from Eq.~\eqref{eq:affinity} is considered as the initial point of $\mathbf{X}$, {\em i.e.}, $\mathbf{X}_0=\mathbf{K}^l_p$. We rewrite Eq.~\eqref{eq:KBQAP} as:
\begin{equation}
\label{eq:obj}
\min_{\mathbf{X}}g(\mathbf{X})=\min_{\mathbf{X}} ||\mathbf{A}_D-\mathbf{X}\mathbf{B}_D\mathbf{X}^T||^2_F-{\rm tr}({\mathbf{X}_u}^T\mathbf{X})
\end{equation}
\begin{algorithm}[t!]
\label{alg1}
\caption{DGM under Quadratic Constraint}
\KwIn{Nodes of graph pairs $V_s$;
two input images $I_s$, where $s$=A, B; the ground truth $\mathbf{X}^*$; initial parameters $\mathbf{W}=\{\mathbf{W}^l_r,\mathbf{W}^l_s,\mathbf{W}^l_{aff}\}$}
\KwOut{prediction $\mathbf{X}_P\in\mathcal{P}$}

    //feature extraction and alignment\\
    $\mathbf{U}_s ~\gets \text{align}(\text{CNN}_{{\text{relu4}}\_2}(I_s), V_s)$;\\
    $\mathbf{F}_s ~\gets \text{align}(\text{CNN}_{{\text{relu5}}\_1}(I_s), V_s)$;\\
    //node attributes\\
    $\mathbf{P}_s$ $\gets$ ${\rm cat}(\mathbf{U}_s;\mathbf{F}_s;[\hat{\mathbf{x}}_s,\hat{\mathbf{y}}_s])$;\\
    $\textbf{Training stage}:$\\
    \For{{\rm epoch $k\leq n$}}{
    $\mathbf{P}^{l}_s$ $\gets$ ${\rm GCN}((\mathbf{A}_D)_s,\mathbf{P}^{l-1}_s)$;\\
    ($\mathbf{A}^{l}_D)_s$ $\gets$ $f(\mathbf{P}^l_s)\odot \mathbf{A}_s$;\\
    $\mathbf{K}^l_p$ $\gets$ ${\rm exp}\{\mathbf{P}^l_A\mathbf{W}^l_{{\rm aff}}{\mathbf{P}^l_B}^T\}$;\\
    $\mathbf{X}$ $\gets$ $\mathbf{K}^l_p$;\\
    \For{{\rm iter = 1:$m_1$}}{
     \For{{\rm k = 1:$m_2$}}{$\noindent$ $\mathbf{y}$ $\gets$ $\arg\min\limits_{\mathbf{y}}\nabla g(\mathbf{X})^T\mathbf{y}$;\\
    //$f_{\mathcal{D}}$ is Sinkhorn normalization\\
    $\mathbf{s}$ $\gets$ $f_{\mathcal{D}}(\mathbf{y})$;\\
    $\overline{\mathbf{X}}$ $\gets$ $\mathbf{X} - \epsilon_k(\mathbf{X} - \mathbf{s})$;
    }
    $\mathbf{X}$ $\gets$ $f_{\mathcal{D}}(\overline{\mathbf{X}})$;
    }
    $\mathbf{W}$ $\gets$ $L_{fm}$($\mathbf{X}$, $\mathbf{X}^{\rm GT}$), $\mathbf{W}$;
    }
    $\textbf{Inference stage}:$\\
    $\mathbf{X}_P$ $\gets$ $\mathbf{X}$;\\
    \For{{\rm iter = 1:m}}{
    \Repeat {{\rm $\mathbf{X}_P$ converges}}{$\noindent$ $\mathbf{y}$ $\gets$ $\arg\min\limits_{\mathbf{y}}\nabla g(\mathbf{X}_P)^T\mathbf{y}$;\\
 //$f_{\mathcal{P}}$ is Hungarian algorithm\\
    $\mathbf{s}$ $\gets$ $f_{\mathcal{P}}(\mathbf{y})$;\\
    $\overline{\mathbf{X}}_{P}$ $\gets$ $\mathbf{X}_P - \epsilon_k(\mathbf{X}_P - \mathbf{s})$;}
    $\mathbf{X}_{P}$ $\gets$ $f_{\mathcal{P}}(\overline{\mathbf{X}}_{P})$;}
\end{algorithm}
We specify unary affinity matrix $\mathbf{X}_u$ as the obtained node affinity matrix $\mathbf{K}^l_p$ in both the training stage and the inference stage.

Due to the paradox between the combinatorial nature of QAP formulation and the differentiable requirement of deep learning framework, we consider a relaxed version of Eq.~\eqref{eq:obj}: $\mathbf{X}\in[0,1]^{n_A\times n_B}, \mathbf{X1}_{n_B}=1, \mathbf{X}^T\mathbf{1}_{n_A}\leq\mathbf{1}_{n_A}$. The value of $\mathbf{X}$ is continuous and satisfies normalization constraints at the same time. By minimizing the objective function, the solution will close to the ground truth in the direction of minimizing the adjacency inconsistency.

\subsection{Quadratic constrained-optimization}
We adopt a differentiable Frank-Wolfe algorithm~\cite{lacoste2015global} for $g(\mathbf{X})$ to obtain an approximate solution:
\begin{align}
\label{eq:yk}
\mathbf y_k &= \arg \min_{\mathbf y} \nabla g(\mathbf{X}_k)^T \mathbf y \\
\label{eq:s}
\mathbf s &= f_{proj}(\mathbf y_k) \\
\label{eq:xhat}
\bar{\mathbf X}_{k+1} &= \mathbf{X}_k - \epsilon_k(\mathbf{X}_k - \mathbf s) \\
\label{eq:x}
\mathbf{X}_{k+1} &= f_{proj}(\bar{\mathbf{X}}_{k+1})
\end{align}
where $\epsilon_k$ is a parameter representing the step size. Usually, we set $\epsilon_k=\frac {2} {k+2}$~\cite{jaggi2013revisiting} in implementation. In training stage, $f_{proj}$ in Eq.~\eqref{eq:s} and Eq.~\eqref{eq:x} is the Sinkhorn layer to project the positive variable into the set of doubly stochastic matrices $\mathcal{D}$, while $f_{proj}$ is the Hungarian algorithm in the inference stage for obtaining a discrete solution. We write down the gradient of $g(\mathbf{X})$ as:
\begin{equation}
\label{eq:gradient}
\nabla g\left(\mathbf{X}\right)=-2[\mathbf{U}^T\mathbf{XB}^l_D+\mathbf{U}{\mathbf{XB}^l_D}^T]-\mathbf{X}_u
\end{equation}
where $\mathbf{U}=\mathbf{A}^l_D-\mathbf{XB}^l_D\mathbf{X}^T$. In training stage, the variable $\mathbf{X}$ is associated with the learnable parameters $\mathbf{W}=\{\mathbf{W}^l_r,\mathbf{W}^l_s,\mathbf{W}^l_{aff}\}$ and every iteration of Frank-Wolfe algorithm is actually going with the parameters fixed before backpropagation~\cite{ionescu2015training}. From Eq.~\eqref{eq:xhat} and Eq.~\eqref{eq:x}, $\mathbf{X}_{k+1}$ is differentiable with respect to $\mathbf{X}_{k}$ so $\mathbf{X}_{k+1}$ is differentiable with respect to $\mathbf{X}_0=\mathbf{K}^l_{p}$ by the recursive relations:
\begin{align}
\label{eq:iter}
\mathbf{X}_{k+1}=f_r(\mathbf{X}_k)=f_r\circ f_r(\mathbf{X}_{k-1})=...=f_r^k(\mathbf{X}_0)
\end{align}
where $f_r(\mathbf{X})\triangleq f_{proj}(\mathbf{X}-\epsilon_k(\mathbf{X}-\mathbf{s}))$ is a differentiable function {\em w.r.t.} $\mathbf{X}$ in training stage.
Since the goal of the optimization is to utilize the information of the pairwise term in Eq.~\eqref{eq:obj}, only few iterations roughly approaching to a local minimum in training stage can fulfill our purpose. Though there is no guarantee of global minimum, it is actually not necessary because the pairwise term will be noisy with outliers and the global minimum may not be the desired matching. Besides, few iterations in training stage encourage the model to learn a relatively short path to approximate the object so that it is easier to convergence within fewer iterations in inference stage.

\iffalse
Though deep graph matching framework can extract CNN features for matching rather than handcrafted attributes as in traditional QAP formulation, there are two main reasons why CNN features along are not enough for graph matching. Firstly because the spatial resolution of feature map on high level convolution layer is not the same as input image and the features from CNN layers will only be interpolated to the graph nodes as approximation. Secondly because the nodes with similar local patches have the similar CNN features while their structure context may be very different. Although some works~\cite{fey2020deep,rolinek2020deep,yu2020learning} explicitly consider the learned or handcrafted pairwise features as a complement to the unary node features, the problems about CNN features mentioned above can not be avoid. Besides, the refinement step of using GCN is also a compromise because it is essentially a smooth operation over neighbor features of each node and can incorporate the information of graph structure only to a limit extent. Thus far, none of the above parts explicitly involve the structure constraint to maximize the adjacency consensus or minimize the adjacency inconsistency.
\fi

\subsection{False-matching loss}
\label{subsec:fmloss}
$\textbf{Class imbalance.}$ The elements of correspondence matrix can be divided into two classes with clear different meaning, one of which represents matched pairs and the other represents unmatched pairs. Given two graphs with equal number of nodes $n$, there are $n$ elements of correspondence matrix are 1 while the rest $n^2-n$ are 0. Under the constraint of one-to-one matching, the unmatched pairs clearly take the majority. Though the values of the soft correspondence matrix are all between 0 and 1, the two classes should be separately treated in the calculation of loss during training.

The cross-entropy-type loss functions such as permutation loss~\cite{wang2019learning} and its improved version~\cite{yu2020learning} achieve the state-of-the-art performance. This type of loss functions directly measures the linear assignment cost between the ground truth $\mathbf{X}^*\in\{0,1\}^{n\times n}$ and the prediction $\mathbf{X}\in[0,1]^{n\times n}$ like
\begin{align}
\label{eq:ce}
L_{ce}=-\sum_{i, j}\mathbf{X}^*_{ij} \log \mathbf{X}_{ij}+(1-\mathbf{X}^*_{ij}) \log (1-\mathbf{X}_{ij})
\end{align}
with $i\in\mathcal{G}_A$, $j\in\mathcal{G}_B$. Cross entropy loss does not consider the class imbalance in deep graph matching because the vast majority of $\mathbf{1-X}$ are preserved by multiplying $\mathbf{1-X^*}$ while at most $n$ elements of $\mathbf{X}$ are kept by multiplying $\mathbf{X^*}$. Similarly, cross entropy loss with Hungarian attention~\cite{yu2020learning} can not solve this issue either and the so called Hungarian attention in loss function brings discontinuity in learning.
\begin{figure}[t!]
\begin{center}
   \includegraphics[width=1\linewidth]{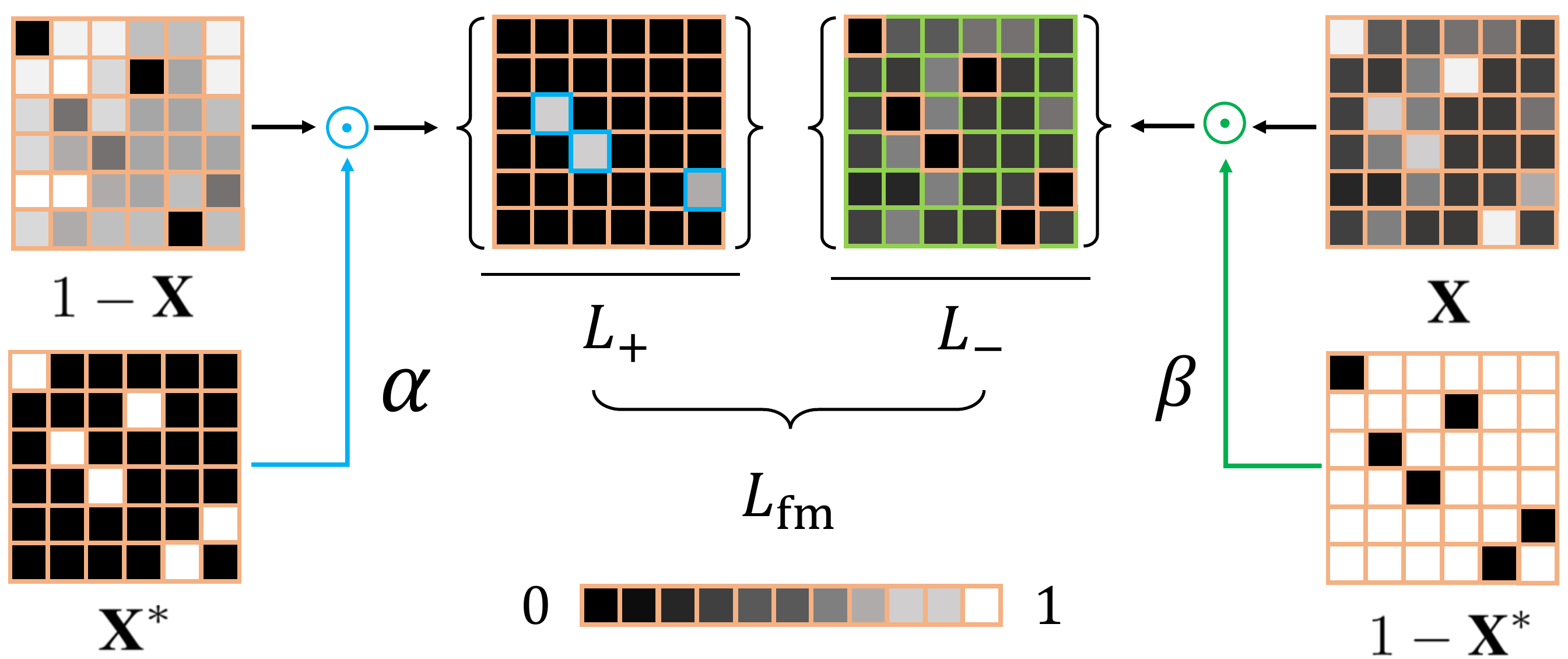}
\end{center}
   \caption{An illustration of the work flow about our proposed false matching loss. We take both the false positive and false negative of the predicted doubly stochastic matrix into consideration. $\alpha$ and $\beta$ are two parameters to weight the two terms.}
\label{fig:inner}
\vspace{-3mm}
\end{figure}

$\textbf{Numerical stability.}$ Since the optimization of deep learning is unconstraint, the local minimum with bad properties can not be avoid. Once a bad case or a bad local minimum occurs, {\em e.g.} at least one element of the prediction $\mathbf{X}$ approaches to 0 while the corresponding element of the ground truth $\mathbf{X}^*$ is 1, the cross-entropy-type loss will be invalid. Though permutation loss with Hungarian attention mechanism is more focus on a specific portion of the prediction $\mathbf{X}$, it only works without severe bad cases either. Even with gradient clipping, overfitting is still a problem to be addressed.
\begin{figure}[t]
\centering
\subfigure[cross entropy loss $L_{ce}$]{
\begin{minipage}[t]{0.5\linewidth}
\centering
\includegraphics[width=1.7in]{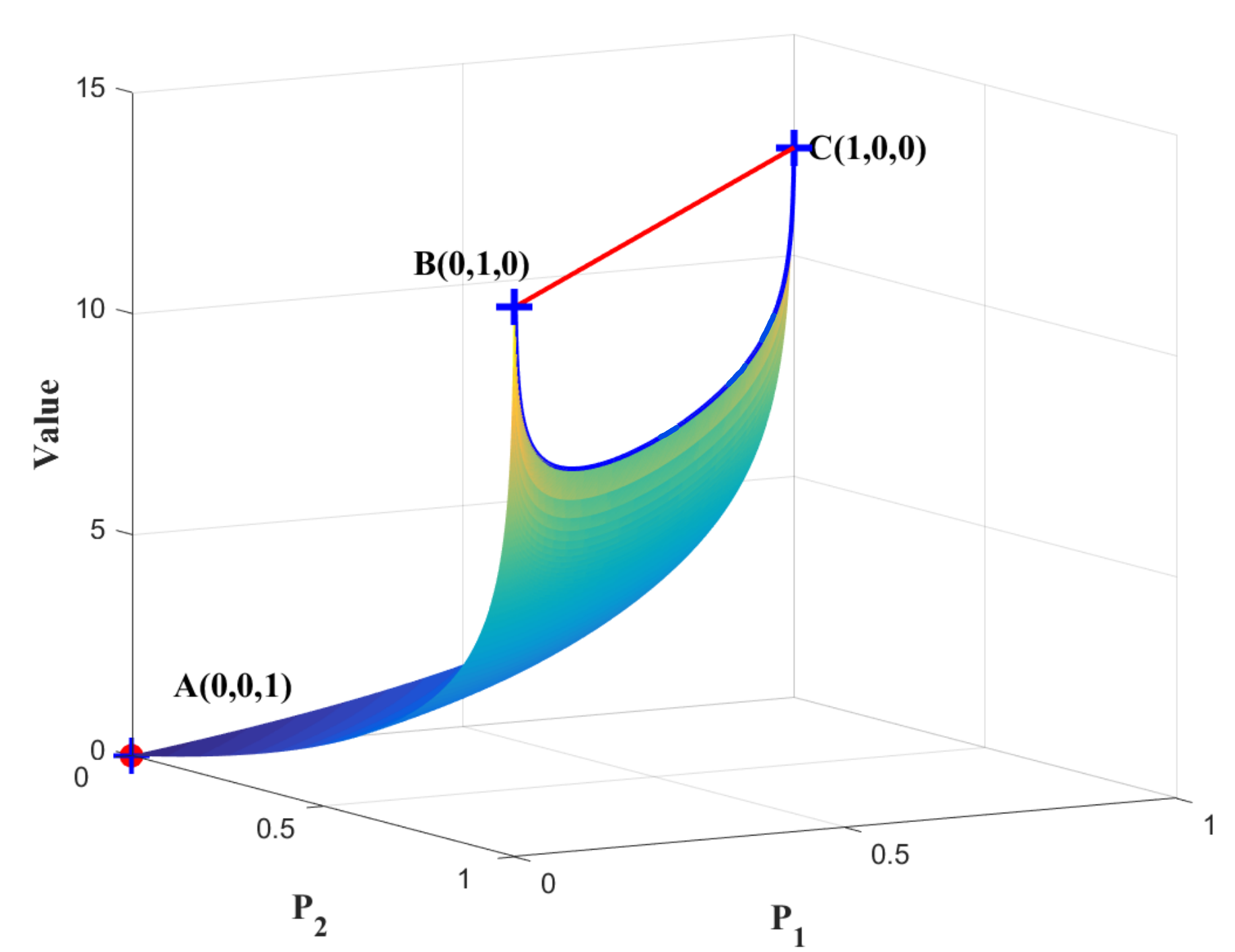}
\end{minipage}%
}%
\subfigure[false matching loss $L_{fm}$]{
\begin{minipage}[t]{0.5\linewidth}
\centering
\includegraphics[width=1.7in]{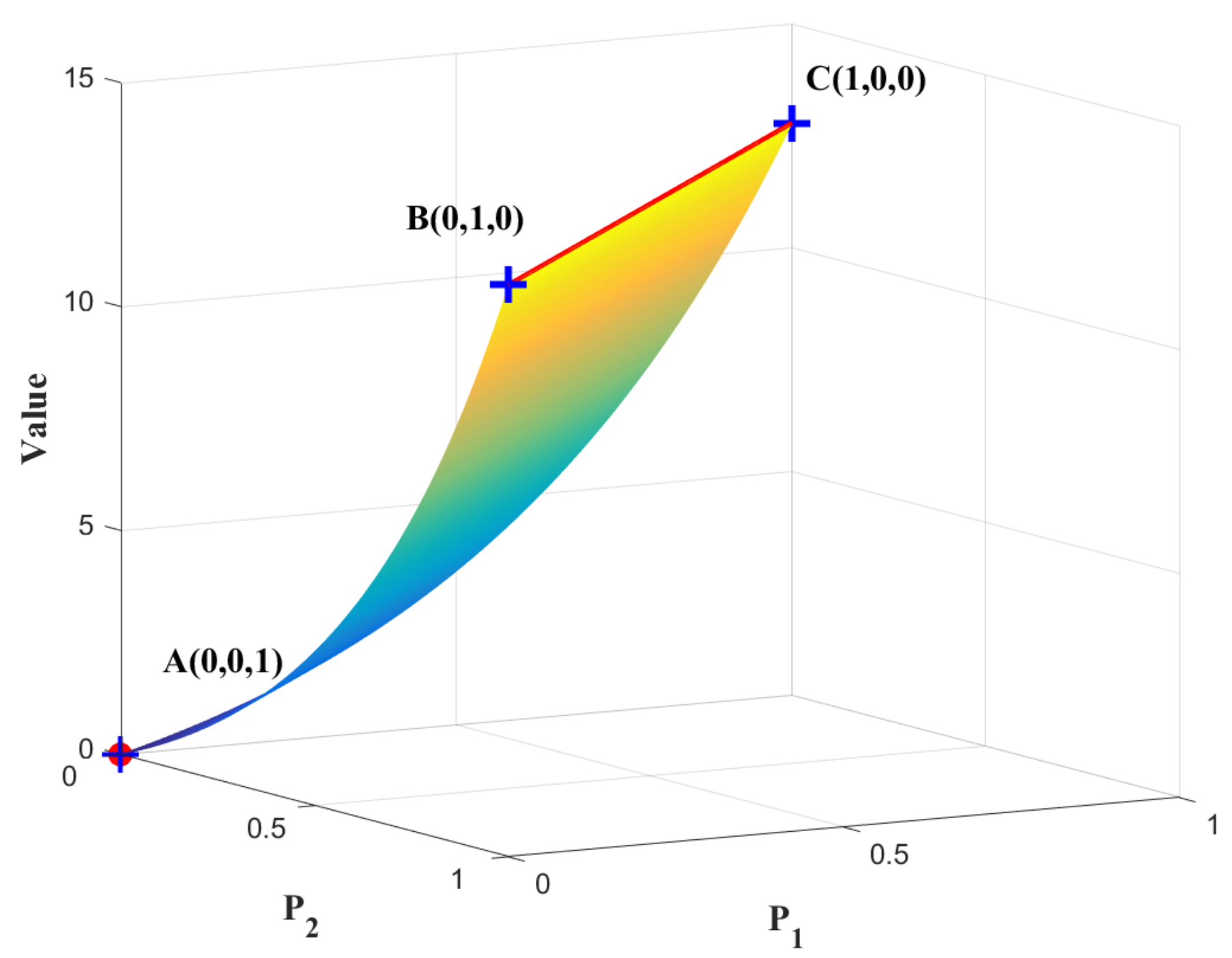}
\end{minipage}%
}%
\centering
\caption{Comparison between cross entropy loss and false matching loss on a toy example, where $\mathbf{X}\in\mathbb{R}^{1\times 3}$ and $\mathbf{X}^*=(0,0,1)$. We mark the three extreme points of $\mathbf{X}$: $\mathbf{A}=(0,0,1)$, $\mathbf{B}=(0,1,0)$ and $\mathbf{C}=(1,0,0)$. A cooler color means a smaller value. The comparison shows that false matching loss has a higher value than cross entropy loss when it close to the ground truth. Besides, false matching loss has a high but limited value when $\mathbf{X}$ close to the extreme points, which are the bad cases for cross entropy loss and make the loss function output infinite values. The curve of cross entropy loss on points between $\mathbf{B}$ and $\mathbf{C}$ are below the red line connected to $\mathbf{B}$ and $\mathbf{C}$, while the curve of false matching loss coincides with the red line.}
\label{fig:innercross}
\vspace{-2mm}
\end{figure}

Facing the problems above, we propose a new loss function called false matching loss:
\begin{align}
\label{eq:fm}
L_{fm}
=\underbrace{e^{\alpha\sum_{i,j}\left[\mathbf{X}\odot(\mathbf{1}-\mathbf{X}^*)\right]_{ij}}}_{L_+} + \underbrace{e^{\beta\sum_{i,j}\left[\mathbf{X}^*\odot(\mathbf{1}-\mathbf{X})\right]_{ij}}}_{L_-}
\end{align}
where $i\in\mathcal{G}_A$, $j\in\mathcal{G}_B $. $L_+$ and $L_-$ of our false matching loss penalize false positive matches and false negative matches, which are the main indexes describing the false matchings in statistics. To address the issue of class imbalance, we draw on the experience of focal loss~\cite{lin2017focal} and take two hyperparameters $\alpha$ and $\beta$ for weighting.

As illustrated in Fig.~\ref{fig:innercross}, false matching loss is more smooth and always has a relatively large value than cross entropy loss even when the prediction $\mathbf{X}$ is close to the ground truth. While on extreme points that are away from the ground truth, false matching loss will have more proper outputs (high but limited). Besides, the values of cross entropy loss on points between $\mathbf{B}$ and $\mathbf{C}$ are lower than that on $\mathbf{B}$ and $\mathbf{C}$, which is not proper because points between $\mathbf{B}$ and $\mathbf{C}$ are still far away from the ground truth.

Our method is summarized in Algorithm~\ref{alg1}.

\section{Experiments and Analysis}
In this section, we evaluate our method on two challenging datasets.
%We implement the proposed method with PyTorch on a Intel Core i7-8700K 3.70GHz CPU with a single Titan XP GPU.
We compare the proposed method with several state-of-the-art deep graph matching methods:  GLMNet~\cite{jiang2019glmnet}, PCA~\cite{wang2019learning},  NGM~\cite{wang2019neural}, ${\rm CIE_1}$~\cite{yu2020learning}, GMN~\cite{zanfir2018deep}, LCSGM~\cite{wang2020learning} and BB-GM~\cite{rolinek2020deep}. To compare the proposed false matching loss with the previous work: permutation loss with Hungarian attention~\cite{yu2020learning}, we follow~\cite{yu2020learning} to implement the loss function for the source code is not publicly available. Given graphs with $n$ nodes, the matching accuracy is computed as
\begin{align}
{\rm accuracy}=\frac{1}{n}\sum_{i,j}[\mathbf{X}^*\odot {\rm Hungarian}(\mathbf{X})]_{ij}
\end{align}

Stochastic gradient descent (SGD)~\cite{bottou2010large} is employed as the optimizer with an initial learning rate $10^{-3}$. In our false matching loss, we set $\alpha=2$ and $\beta=0.1$. In our optimization step, we set the number of iterations $m_1=3$ and $m_2=5$ for the consideration of both computational efficiency and convergence.

Since the feature refinement step is not limited to Eq.~\eqref{eq:GCN}, we additionally deploy another novel architecture SplineCNN~\cite{fey2018splinecnn} to replace Eq.~\eqref{eq:GCN} as a comparision with~\cite{rolinek2020deep}. Thus, two versions of our method are provided: qc-DG${\rm M}_1$ and qc-DG${\rm M}_2$. In qc-DG${\rm M}_1$, the node attributes are refined by a two-layer GCN as Eq.~\eqref{eq:GCN} with ReLU~\cite{nair2010rectified} activation function. While in qc-DG${\rm M}_2$, the refinement step is done by a two-layer SplineCNN~\cite{fey2018splinecnn} with max aggregation. In our implementation, all the input graphs are constructed by Delaunay
triangulation. Specifically for GMN~\cite{zanfir2018deep}, two input graphs are constructed by fully-connected topology and Delaunay triangulation respectively. While ${\rm GMN_D}$ is another version of GMN that two input graphs are both constructed by Delaunay triangulation.

\subsection{Results on Pascal VOC Keypoints}
Pascal VOC dataset~\cite{everingham2010pascal} with Berkeley annotations of keypoints~\cite{bourdev2009poselets} has 20 classes of instance images with annotated keypoints. There are 7,020 annotated images for training and 1,682 for testing. Before training, each instance is cropped around its bounding box and is re-scaled to $256\times 256$ as the input for VGG16. We follow the previous works to filter the truncated, occluded and difficult images. The training process will be performed on all 20 classes. Because of the large variation of pose, scale, and appearance, Pascal VOC Keypoints is considerably a challenging dataset.

\begin{table*}[t!]
\begin{center}
\resizebox{\textwidth}{!}{
\begin{tabular}{c | cccccccccccccccccccc | c}
\toprule
\noalign{\smallskip}
Method & aero & bike & bird & boat & bottle & bus & car & cat & chair & cow & table & dog & horse & mbike & person & plant & sheep & sofa & train & tv & Mean\\
\noalign{\smallskip}
\midrule
\noalign{\smallskip}
  GMN~\cite{zanfir2018deep} & 35.5 & 50.0 & 52.2 & 45.2 & 75.5 & 69.1 & 59.0 & 61.1 & 34.4 & 51.9 & 66.4 & 53.4 & 53.9 & 50.7 & 31.1 & 75.8 & 59.2 & 47.9 & 86.1 & 89.6 & 57.4 \\
  ${\rm GMN_D}$~\cite{zanfir2018deep} & 40.8 & 57.1 & 56.2 & 48.2 & 75.5 & 71.6 & 64.1 & 63.3 & 36.3 & 54.3 & 50.2 & 57.6 & 60.3 & 55.4 & 35.5 & 85.2 & 62.7 & 51.8 & 86.5 & 87.4 & 60.0\\
  qc-GMN & 37.3 & 52.2 & 54.3 & 47.2 & 76.4 & 70.4 & 61.2 & 61.7 & 34.5 & 53.1 & 69.1 & 55.4 & 56.0 & 52.4 & 31.6 & 77.3 & 59.7 & 49.1 & 87.4 & 89.7 &  58.8 \\
  PCA~\cite{wang2019learning} & 40.9 & 55.0 & 65.8 & 47.9 & 76.9 & 77.9 & 63.5 & 67.4 & 33.7 & 66.5 & 63.6 & 61.3 & 58.9 & 62.8 & 44.9 & 77.5 & 67.4 & 57.5 & 86.7 & 90.9 & 63.8 \\
  qc-PCA & 42.5 & 58.5 & 66.1 & 51.3 & 79.6 & 78.2 & 65.8
  & 68.7 & 35.1 & 66.8 & 65.6 & 62.5 & 62.1 &  63.1 &  45.1 & 80.7 &  67.7 &  59.1 & 87.0 & 91.1 &  64.8\\
  PCA-H~\cite{wang2019learning} & 50.8 & 61.7 & 62.6 & 56.4 & 80.0 & 75.6 & 72.4 & 74.0 & 38.5 & 64.3 & 49.9 & 63.8 & 65.2 & 63.5 & 46.0 & 78.5 & 68.0 & 41.5 & 82.2 & 90.8 & 64.3 \\
  PCA-F & 50.0 & 66.7 & 61.8 & 55.1 & 81.5 & 75.5 & 70.1 & 70.4 & 39.7 & 64.8 & 60.3 & 65.5 & 67.6 & 64.2 & 45.6 & 84.4 & 68.6 & 56.5 & 88.7 & 91.1 & 66.3 \\
  NGM~\cite{wang2019neural} & 50.8 & 64.5 & 59.5 & 57.6 & 79.4 & 76.9 & 74.4 & 69.9 & 41.5 & 62.3 & 68.5 & 62.2 & 62.4 & 64.7 & 47.8 & 78.7 & 66.0 & 63.3 & 81.4 & 89.6 & 66.1 \\
  GLMNet~\cite{jiang2019glmnet} & $\bf{52.0}$ & 67.3 & 63.2 & 57.4 & 80.3 & 74.6 & 70.0 & 72.6 & 38.9 & 66.3 & $\bf{77.3}$ & 65.7 & 67.9 & 64.2 & 44.8 & 86.3 & 69.0 & 61.9 & 79.3 & 91.3 & 67.5 \\
  LCSGM~\cite{wang2020learning} & 46.9 & 58.0 & 63.6 & $\bf{69.9}$ & $\bf{87.8}$ & 79.8 & 71.8 & 60.3 & $\bf{44.8}$ & 64.3 & 79.4 & 57.5 & 64.4 & 57.6 & $\bf{52.4}$ & $\bf{96.1}$ & 62.9 & 65.8 & $\bf{94.4}$ & 92.0 & 68.5\\
  ${\rm CIE_1}$-H~\cite{yu2020learning} & 51.2 & $\bf{69.2}$ & $\bf{70.1}$ & 55.0 & 82.8 & 72.8 & 69.0 & $\bf{74.2}$ & 39.6 & 68.8 & 71.8 & $\bf{70.0}$ & $\bf{71.8}$ & 66.8 & 44.8 & 85.2 & $\bf{69.9}$ & 65.4 & 85.2 & $\bf{92.4}$ & 68.9\\
\noalign{\smallskip}
\midrule
\noalign{\smallskip}
 qc-DG${\rm M}_1$(ours) & 48.4 & 61.6 & 65.3 & 61.3 & 82.4 & 79.6 & 74.3 & 72.0 & 41.8 & $\bf{68.8}$ & 65.0 & 66.1 & 70.9 & 69.6 & 48.2 & 92.1 & 69.0 & 66.7 & 90.4 & 91.8 & 69.3 \\
 qc-DG${\rm M}_2$(ours) & 49.6 & 64.6 & 67.1 & 62.4 & 82.1 & $\bf{79.9}$ & $\bf{74.8}$ & 73.5 & 43.0 & 68.4 & 66.5 & 67.2 & 71.4 & $\bf{70.1}$ & 48.6 & 92.4 & 69.2 & $\bf{70.9}$ & 90.9 & 92.0 & $\bf{70.3}$ \\
\noalign{\smallskip}
\bottomrule
\end{tabular}}
\end{center}
\vspace{-3mm}
\caption{Accuracy (\%) of 20 classes and average accuracy on Pascal VOC. Bold numbers represent the best performing of the methods to be compared. Method with ``-H" and ``-F" denotes it with permutation loss with Hungarian attention and false matching loss, respectively. Suffix ``-QC" denotes the method with the proposed quadratic constraint.}
\label{table:accuracyP}
\end{table*}

\begin{table*}[t!]
\centering
\resizebox{\textwidth}{!}{
\begin{tabular}{c | cccccccccccccccccccc | c}
    \toprule
    \noalign{\smallskip}
    Method & aero & bike & bird & boat & bottle & bus & car & cat & chair & cow & table & dog & horse & mbike & person & plant & sheep & sofa & train & tv & Mean\\
    \noalign{\smallskip}
    \midrule
    \noalign{\smallskip}
      BB-GM-Max~\cite{rolinek2020deep} & 35.5 & 68.6 & 46.7 & 36.1 & 85.4 & 58.1 & 25.6 & 51.7 & 27.3 & 51.0 & 46.0 & 46.7 & 48.9 & 58.9 & 29.6 & 93.6 & 42.6 & 35.3 & 70.7 & 79.5 & 51.9\\
     BB-GM~\cite{rolinek2020deep} & 42.7 & 70.9 & 57.5 & 46.6 & 85.8 & 64.1 & 51.0 & 63.8 & 42.4 & 63.7 & 47.9 & 61.5 & 63.4 & 69.0 & 46.1 & 94.2 & 57.4 & 39.0 & 78.0 & 82.7 & 61.4\\
     \noalign{\smallskip}
     \midrule
    \noalign{\smallskip}
     qc-DGM$_1$ (ours) & 30.1 & 59.1 & 48.6 & 40.0 & 79.7 & 51.6 & 32.4 & 55.4 & 26.1 & 52.1 & 47.0 & 50.1 & 56.8 & 59.9 & 27.6 & 90.4 & 50.9 & 33.1 & 71.3 & 78.8 & 52.0\\
     qc-DGM$_2$ (ours) & 30.9 & 59.8 & 48.8 & 40.5 & 79.6 & 51.7 & 32.5 & 55.8 & 27.5 & 52.1 & 48.0 & 50.7 & 57.3 & 60.3 & 28.1 & 90.8 & 51.0 & 35.5 & 71.5 & 79.9 & 52.6\\
    \noalign{\smallskip}
    \bottomrule
\end{tabular}
}
\vspace{-1mm}
   \caption{F1 score (\%) of matching and mean over 20 classes on Pascal VOC.}
   \label{table:accuracyP1}
    \vspace{-2mm}
\end{table*}

Experimental results on 20 classes are given in Table~\ref{table:accuracyP}. Since the peer work~\cite{rolinek2020deep} suggests to compare F1 score (the harmonic mean of precision and recall) on the dataset without intersection filtering, we provide F1 score of our method to compare with BB-GM~\cite{rolinek2020deep} and its ablation BB-GM-Max as shown in Table~\ref{table:accuracyP1}.

\begin{figure}[t!]
\vspace{-7mm}
\centering
\includegraphics[height=0.13\linewidth, width=0.8\linewidth]{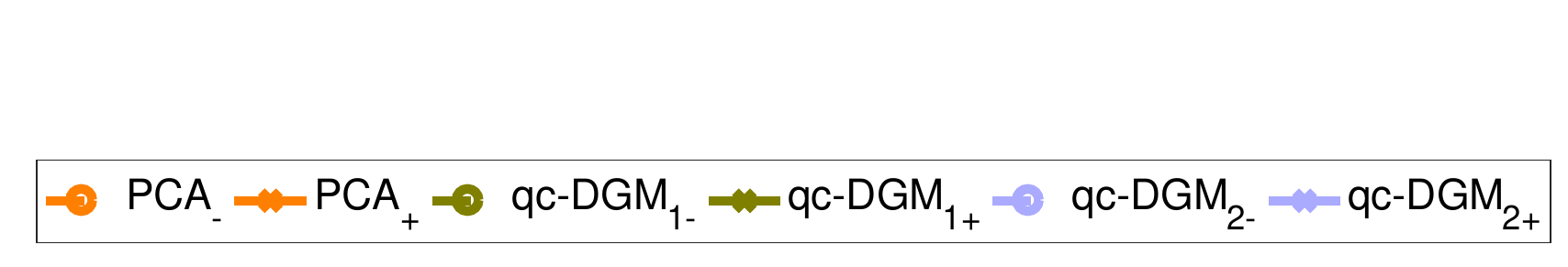}\\
\vspace{-1mm}
\subfigure[]{
\includegraphics[height=0.4\linewidth,width=0.47\linewidth]{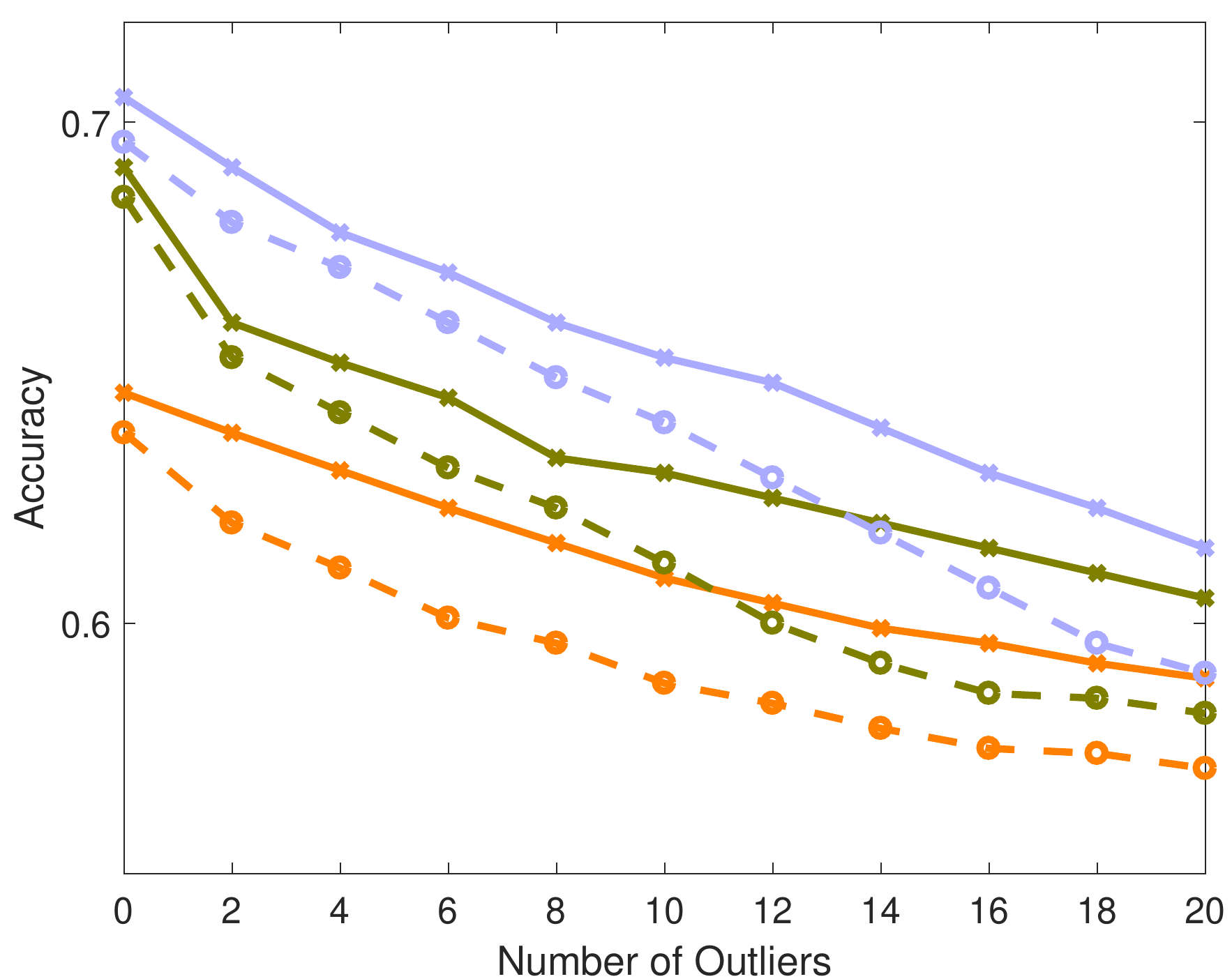}}
\subfigure[]{
\includegraphics[height=0.4\linewidth,width=0.47\linewidth]{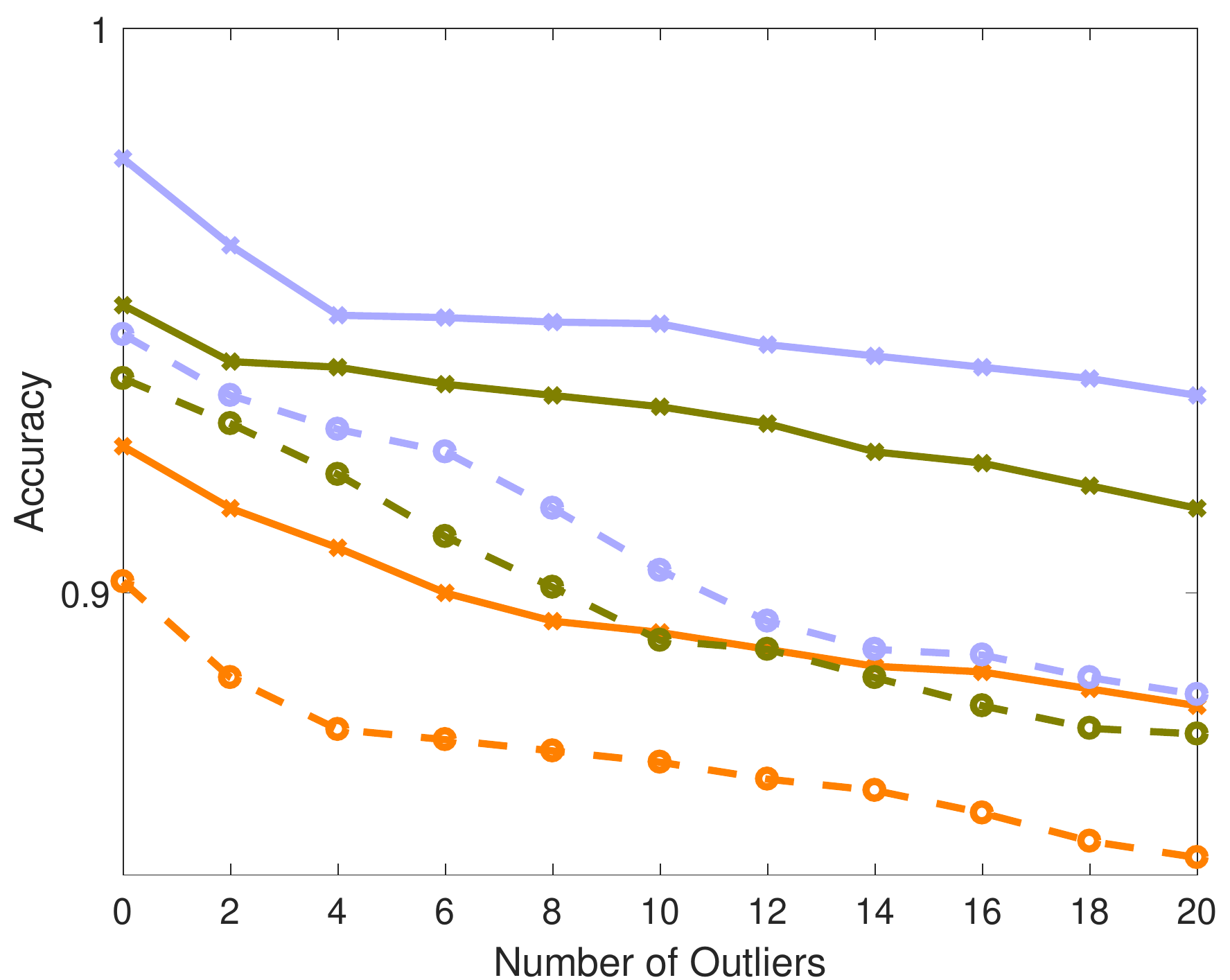}
}
\caption{Robustness analysis against outliers on Pascal VOC (a) and Willow Object Class (b). Method with ``+'' means {\em with} quadratic constraint while ``-'' means {\em without} quadratic constraint.}
\label{fig:comparison}
\vspace{-3mm}
\end{figure}

To show the robustness of the deep graph matching models against outliers, we add outliers to the original set of keypoints. The 2D Cartesian coordinates of the outliers are generated by Gaussian distribution $\mathcal{N}(0, 10)$. The outliers exist only in inference stage to challenge the model trained on the clean data (without outliers). This robustness test is more challenging than that on synthetic graphs because all the inliers and outliers have their extracted CNN features and thus, more close to the real-world scenes. Experimental results are shown in Fig.~\ref{fig:comparison}. With our quadratic constraint, the improvement of overall accuracy is witnessed on the clean data and the robustness of deep graph matching models have been significantly improved against outliers.

By comparing PCA and qc-PCA or GMN and qc-GMN in Table~\ref{table:accuracyP}, the effectiveness of our quadratic constraint is shown to be general and the matching accuracy is improved over all 20 classes by considering our quadratic constraint.

\begin{table}[t!]
\vspace{-1mm}
\scriptsize
\centering
\begin{tabular}{c|ccccc|c}
\noalign{\smallskip}
\toprule
%\noalign{\smallskip}
{Method} & {face} & {mbike} & {car} & {duck} & {wbottle} & {Mean} \\
%\noalign{\smallskip}
\midrule
\noalign{\smallskip}
 HARG-SSVM~\cite{cho2013learning} & 91.2 & 44.4 & 58.4 & 55.2 & 66.6 & 63.2\\
 GMN~\cite{zanfir2018deep} & 98.1 & 65.0 & 72.9 & 74.3 & 70.5 & 76.2 \\
 ${\rm GMN_D}$ & $\textbf{100.0}$ & 82.4 & 84.4 & 84.0 & 91.2 & 88.4 \\
 qc-GMN & $\textbf{100.0}$ & 65.7 & 75.3 & 81.1 & 90.5 & 82.5\\
 PCA~\cite{wang2019learning} & $\textbf{100.0}$ & 76.7 & 84.0 & 93.5 & 96.9 & 90.2 \\
 qc-PCA & $\textbf{100.0}$ & 83.3 & 87.3 & 93.8 & 97.1 & 92.3 \\
 PCA-H~\cite{wang2019learning} & $\textbf{100.0}$ & 76.9 & 88.9 & 89.7 & 92.9 & 89.7 \\
 PCA-F~\cite{wang2019learning} & $\textbf{100.0}$ & 78.4 & 86.8 & 93.2 & 97.2 & 91.1 \\
 NGM~\cite{wang2019neural} & 99.2 & 82.1 & 84.1 & 77.4 & 93.5 & 87.2 \\
 GLMNet~\cite{jiang2019glmnet} & $\textbf{100.0}$ & 89.7 & 93.6 & 85.4 & 93.4 & 92.4 \\
 LCSGM~\cite{wang2020learning} & $\textbf{100.0}$ & $\textbf{99.4}$ & 91.2 & 86.2 & 97.9 & 94.9\\
 ${\rm CIE_1}$-H~\cite{yu2020learning} & $\textbf{100.0}$ & 90.0 & 82.2 & 81.2 & 97.6 & 90.2 \\
 BB-GM~\cite{rolinek2020deep} & $\textbf{100.0}$ & $\textbf{99.2}$ & 96.9 & 89.0 & 98.8 & 96.8\\
\midrule
 qc-DG${\rm M}_1$(ours) &$\textbf{100.0}$ & 95.0 & 93.8 & $\textbf{93.8}$ & 97.6 & 96.0 \\
 qc-DG${\rm M}_2$(ours) & $\textbf{100.0}$ & 98.8 & $\textbf{98.0}$ & 92.8 & $\textbf{99.0}$ & $\textbf{97.7}$\\
\bottomrule
\end{tabular}
\vspace{.2em}
\caption{Accuracy (\%) of 5 classes and the average on Willow Object Class.}
\label{table:willow}
\vspace{-3mm}
\end{table}

\begin{figure*}[t!]
\centering
\includegraphics[height=0.14\linewidth,width=0.33\linewidth]{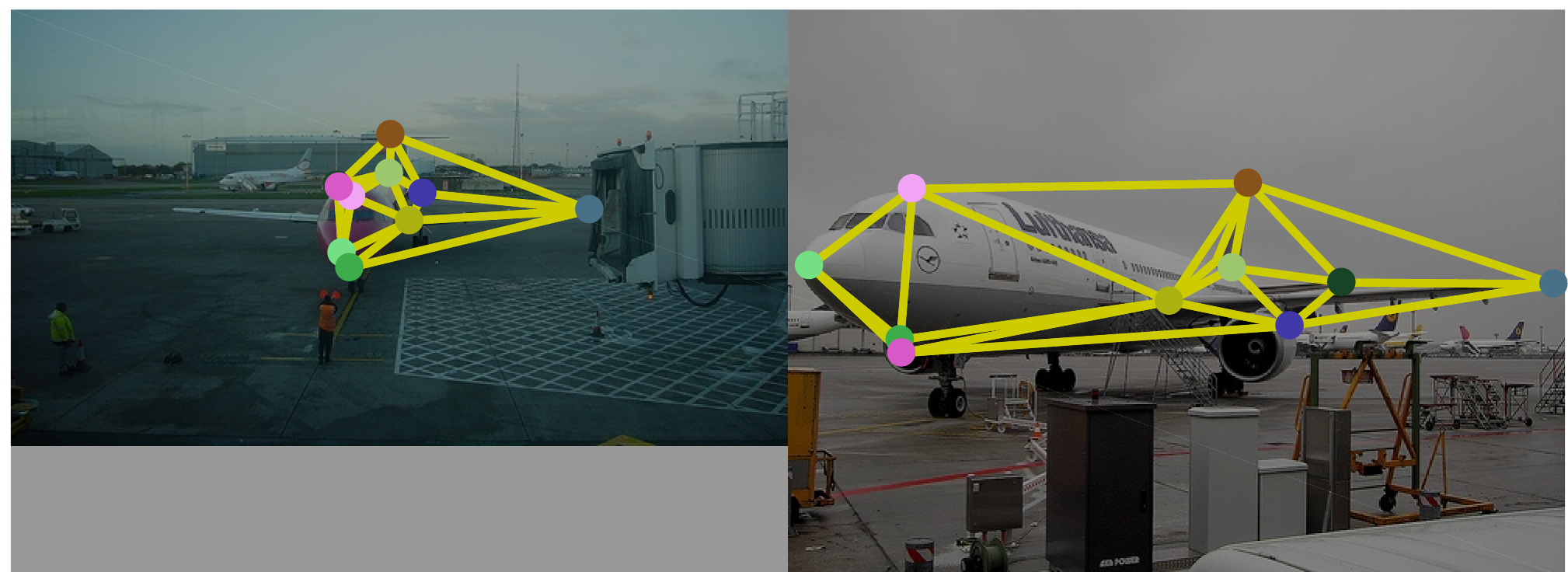}
\includegraphics[height=0.14\linewidth,width=0.33\linewidth]{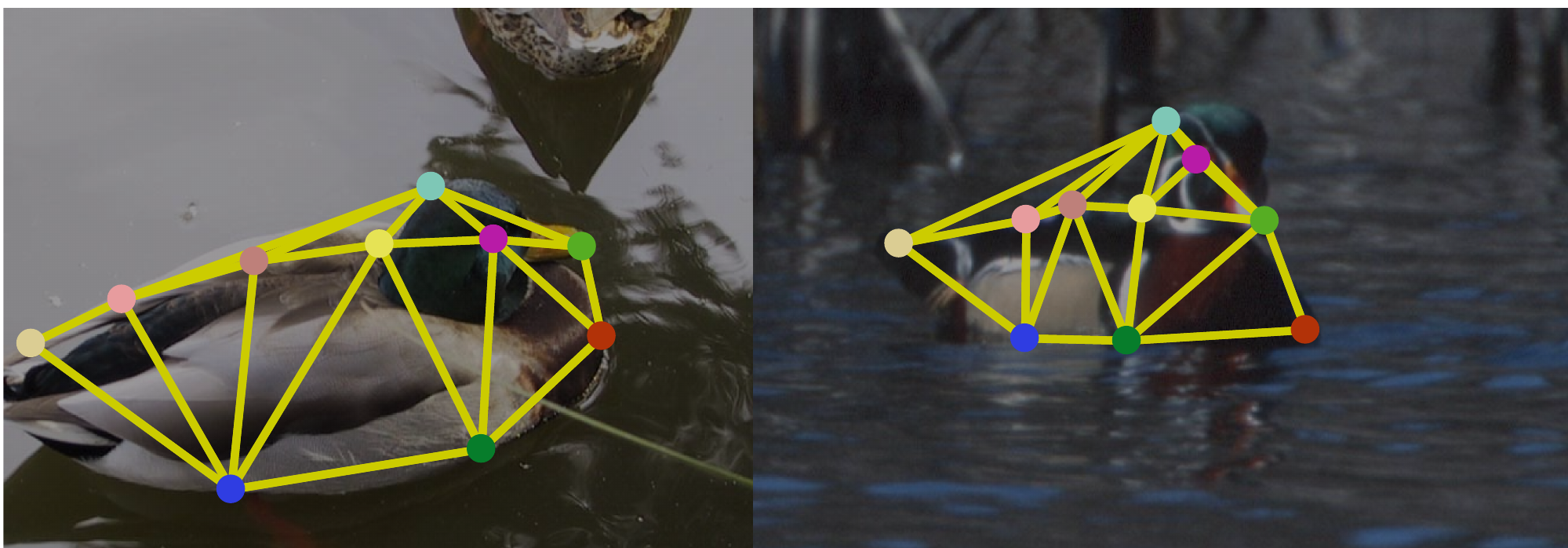}
\includegraphics[height=0.14\linewidth, width=0.33\linewidth]{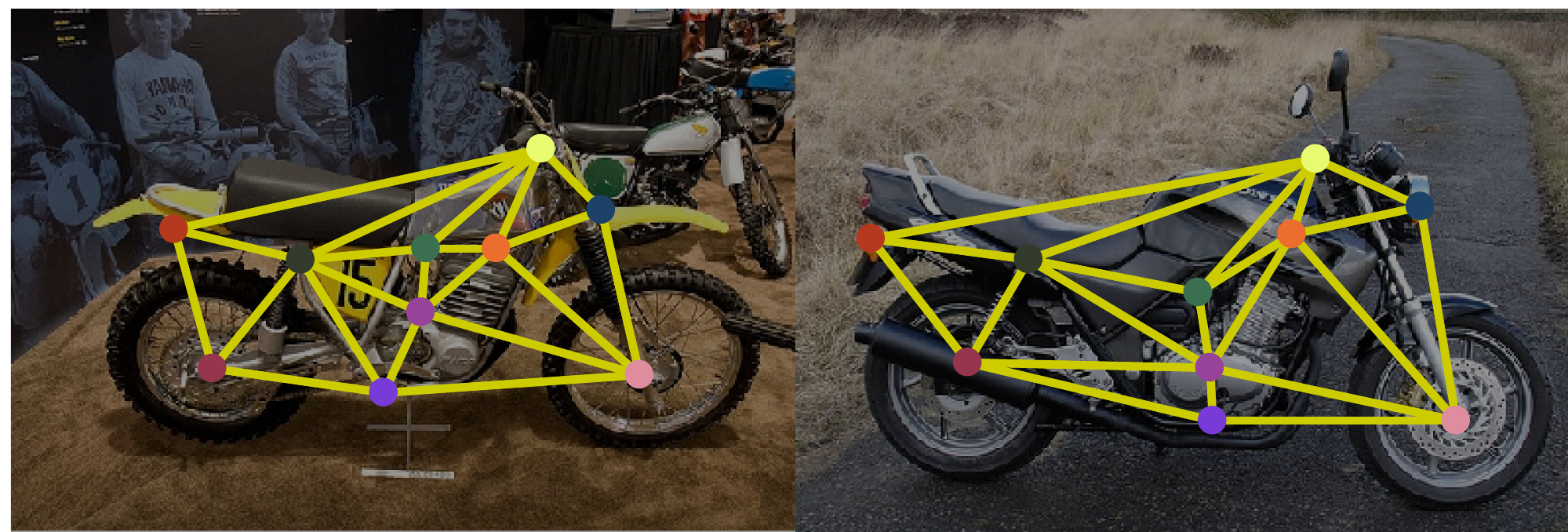}\\
\includegraphics[height=0.14\linewidth,width=0.33\linewidth]{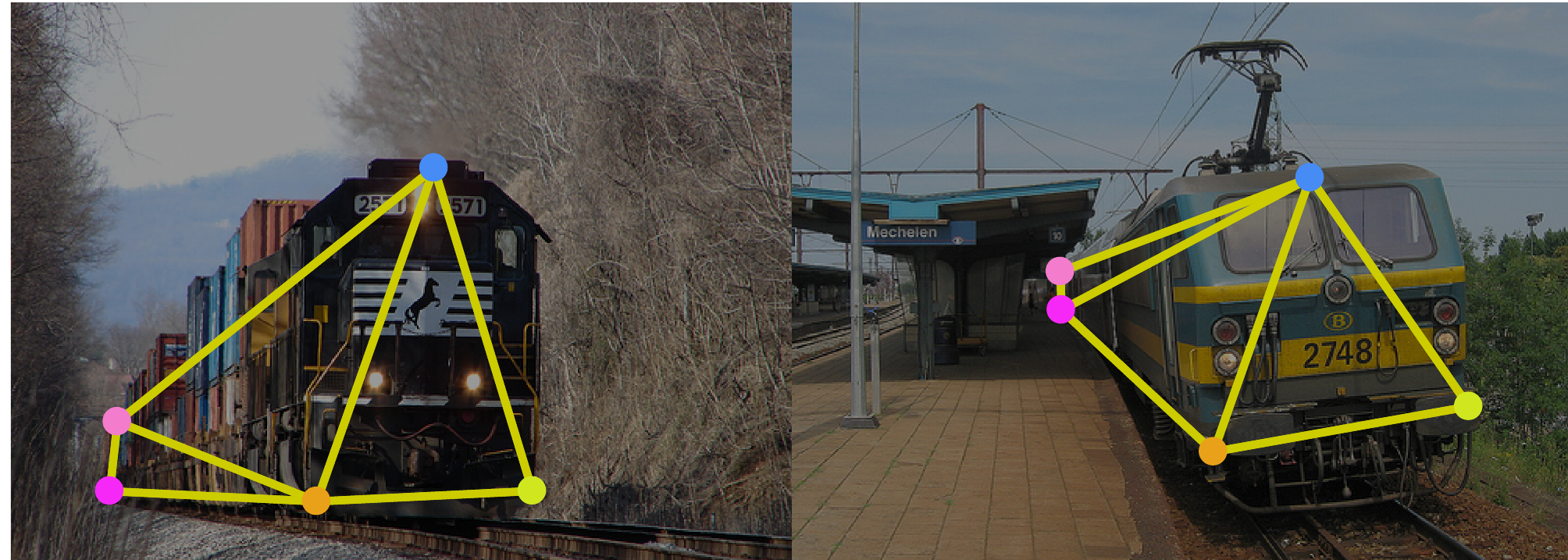}
\includegraphics[height=0.143\linewidth,width=0.33\linewidth]{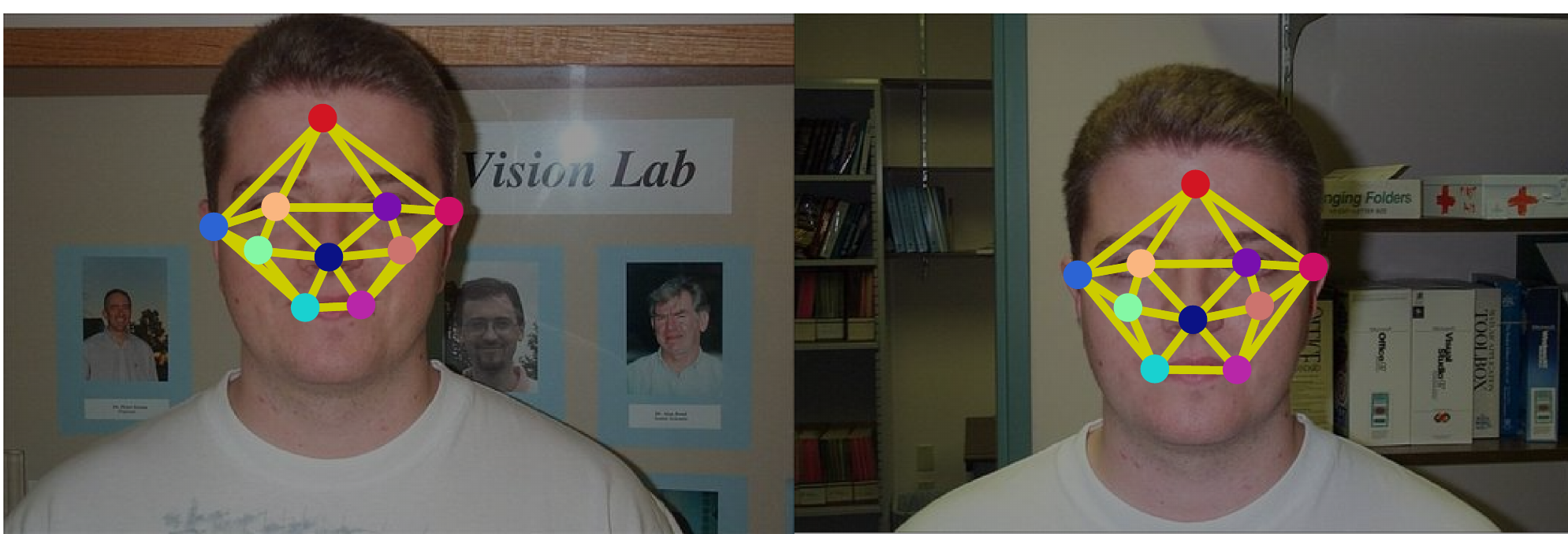}
\includegraphics[height=0.14\linewidth,width=0.33\linewidth]{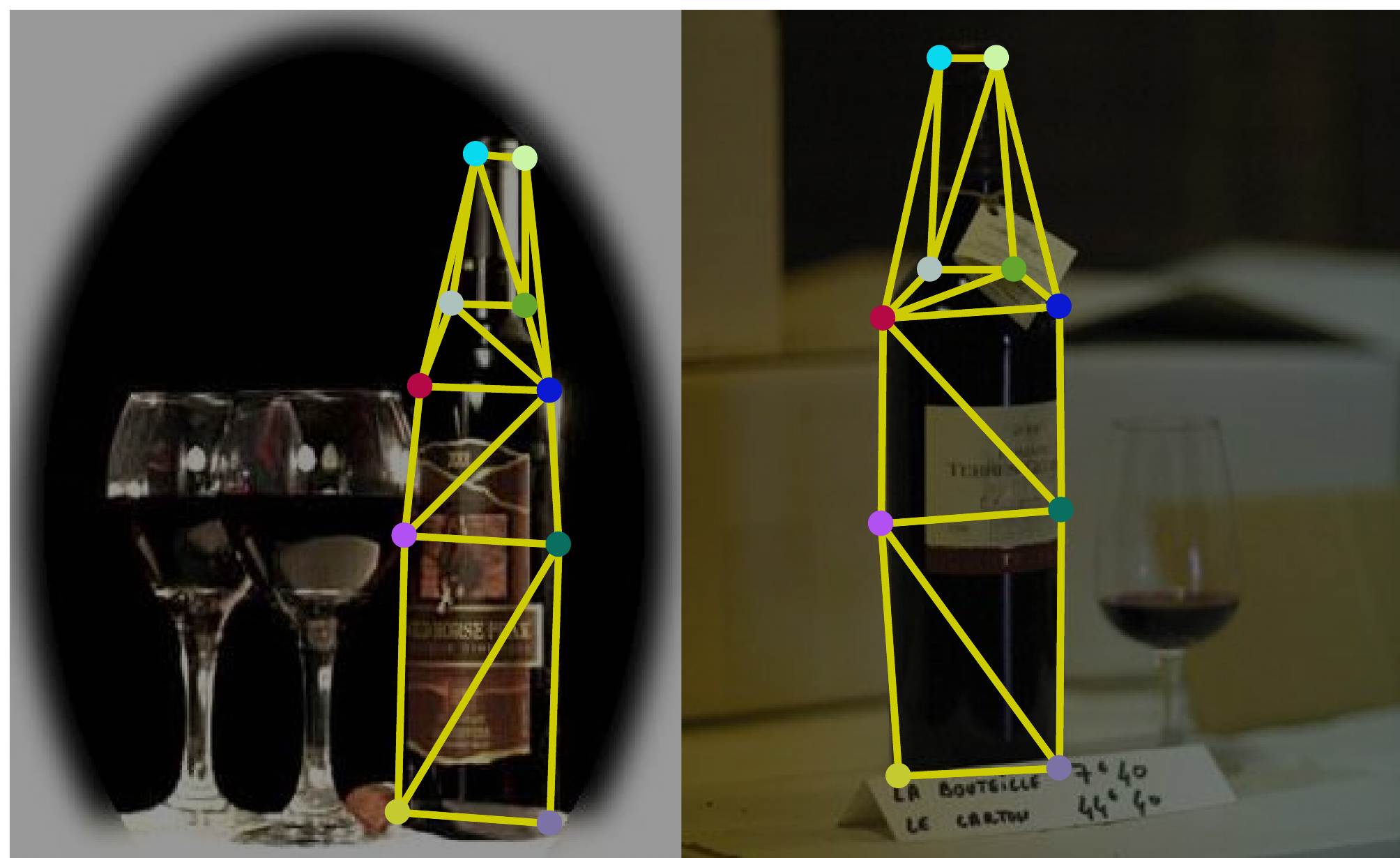}
\caption{Matching examples on Pascal VOC and Willow Object Class. Nodes with the same color indicate the correspondence of a graph pair. All the visualized graphs are constructed by Delaunay triangulation in yellow.}
\vspace{-3mm}
\label{fig:example}
\end{figure*}

\subsection{Results on Willow Object Class}
Willow Object Class dataset~\cite{cho2013learning} contains 5 classes with 256 images in total. Three classes (face, duck and wine bottle) of the dataset are from Caltech-256 and the rest two classes (car and motorbike) are from Pascal VOC 2007. We resize all the image pairs to 256$\times$256 for VGG16 backbone and crop the images around the bounding box of the objects. The variations of the pose, scale, background and illumination of the images on Willow Object Class are small, and thus, graph matching tasks on this dataset are much more easier.

As shown in Table~\ref{table:willow}, the proposed method achieves the state-of-the-art performance. Comparing the methods with and without quadratic constraint (PCA and qc-PCA or GMN and qc-GMN), the performance improvements on Willow dataset are more prominent than those on Pascal VOC, which because the structure variations of graph pairs on the easier dataset are relatively small and quadratic constraint contributes to matching more soundly. Since BB-GM adopts SplineCNN to refine the features, we provide qc-DG${\rm M}_2$ for fair comparison, which has a higher overall accuracy.

\subsection{Further study}
Table~\ref{table:ablation} shows the usefulness of the proposed components. CNN feature vector of each node has d=1024 dimensions, which are supposed to be better than the two-dimensional coordinates but the geometric prior encoded in coordinates provides a more precise description of graph nodes. By simply combining the unary geometric prior with the extracted CNN features, the matching accuracy is improved, which supports our point of view, {\em i.e.}, CNN features are indeed useful but not discriminative enough to depict pixel-wise graph nodes.

$\textbf{Quadratic constraint.}$ There are various forms of quadratic constraint that are not limited to ours. Global affinity matrix $\mathbf{K}$ is constrainted by graph incidence matrices in the factorized form of Lawler's QAP~\cite{lawler1963quadratic,zhou2015factorized}, which can be considered as another form of quadratic constraint adopted in GMN. As shown in Table~\ref{table:accuracyP} and Table~\ref{table:willow}, GMN performs significantly worse than ${\rm GMN_D}$ on both two datasets by replacing similar graph topology with the completely different one, while the extracted deep features of both settings remain the same. This implies the main limitation of quadratic constraint, {\em i.e.}, only graphs with similar topology contribute to matching. Besides, the fact shows the independent relationship between the graph structure and raw deep features for the change of graph topology can not be reflected by raw deep features (before being refined).

\begin{figure}[t!]
\vspace{-1mm}
\centering
  \includegraphics[height=0.13\linewidth, width=0.45\linewidth]{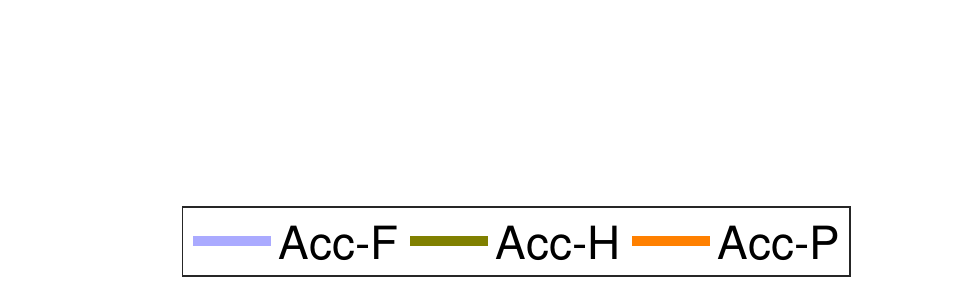}
  %\vspace{2mm}
  \includegraphics[height=0.13\linewidth, width=0.47\linewidth]{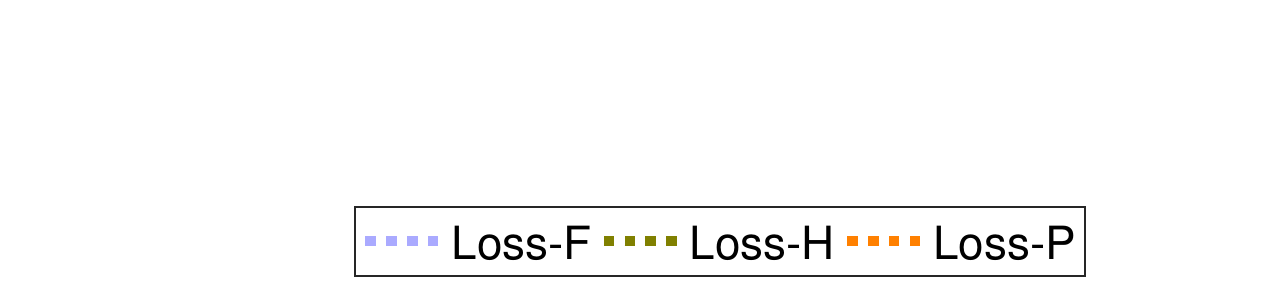}\\
   \includegraphics[height=0.4\linewidth, width=0.47\linewidth]{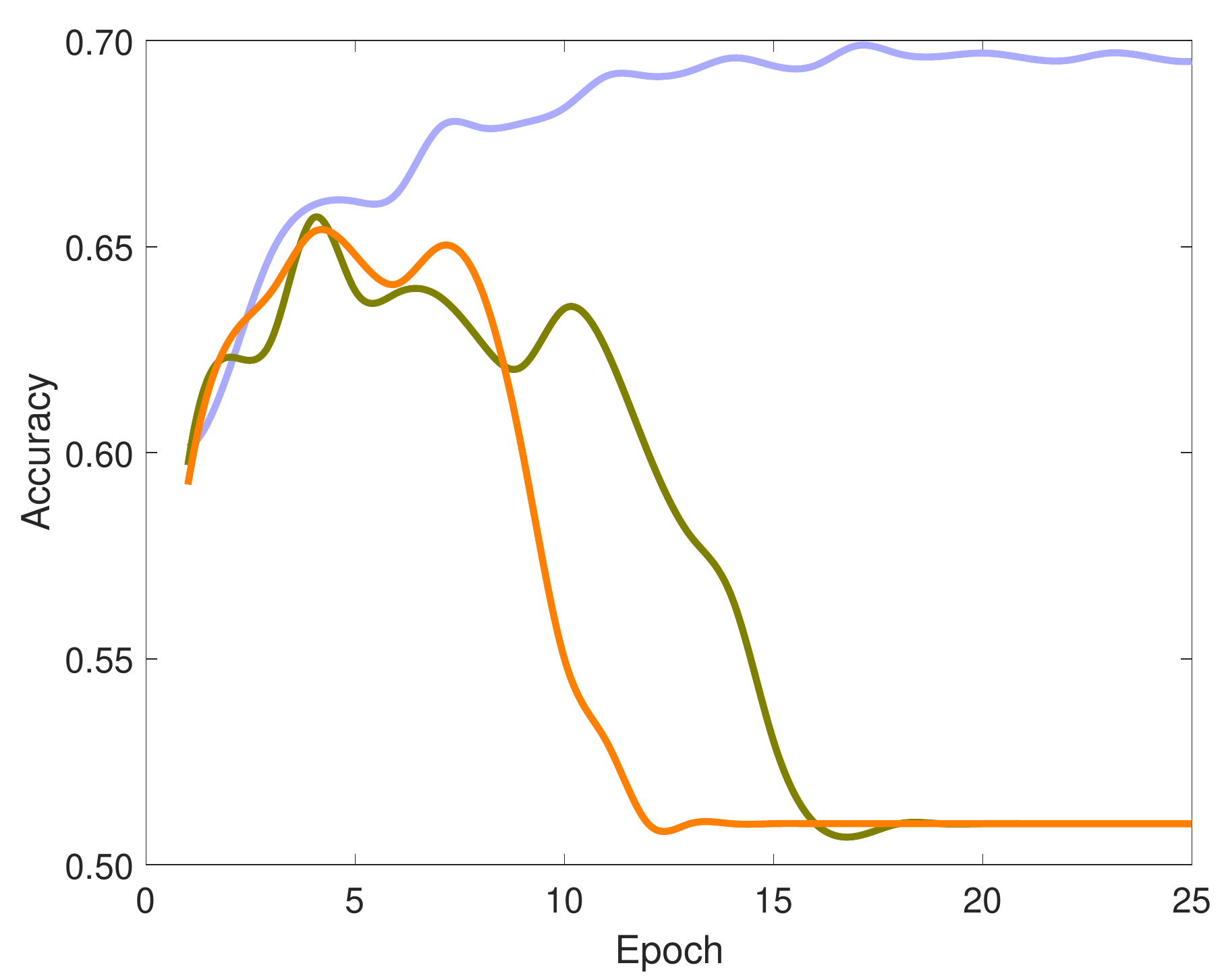}
   \includegraphics[height=0.4\linewidth, width=0.47\linewidth]{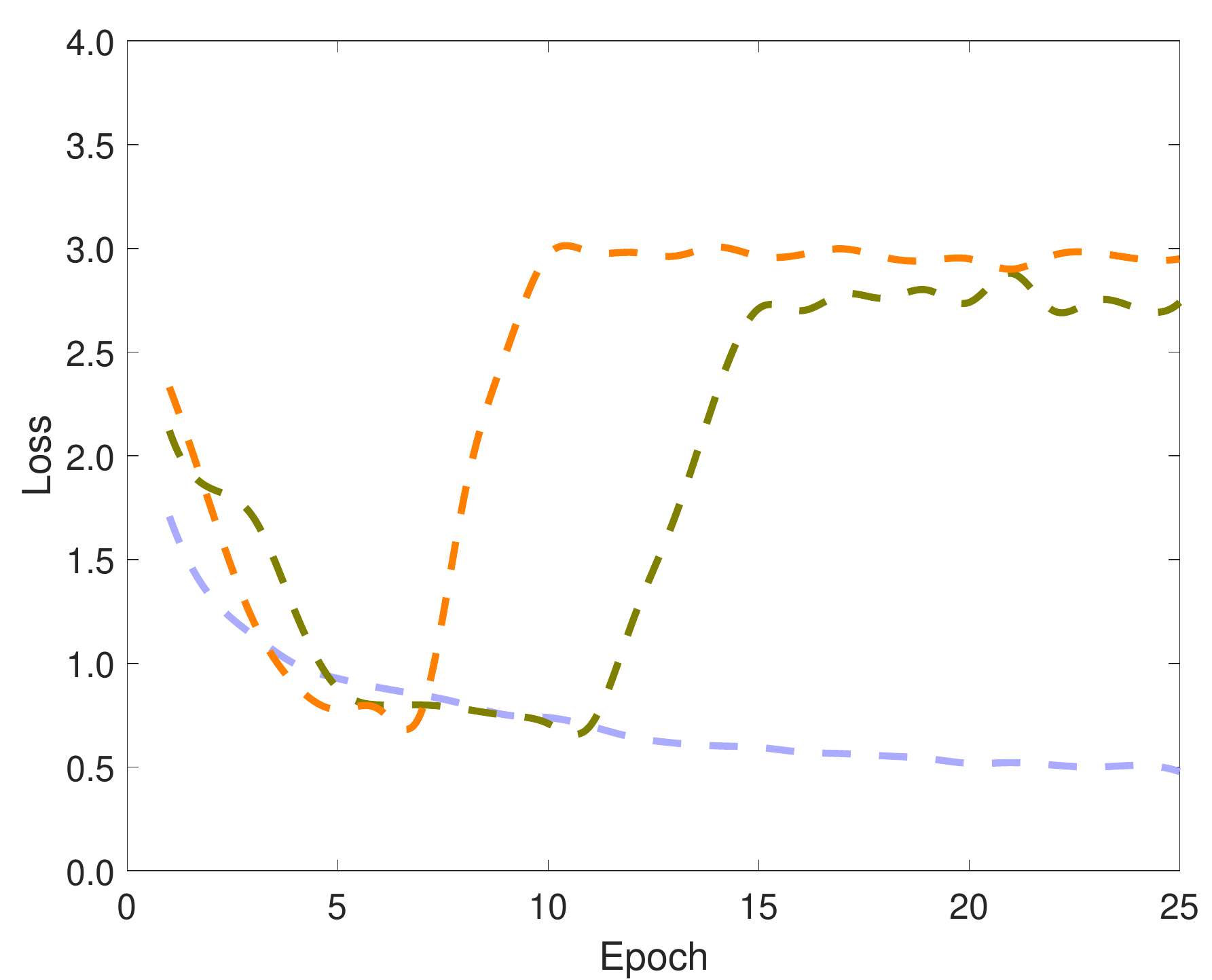}
   %\vspace{-1mm}
   \caption{Accuracy/loss vs. training epoch on Pascal VOC. As the training goes, the loss functions try to drag the output to binary and the local minimum with bad properties makes the cross-entropy-type loss explosion. Since the accuracy will be very close to 0 after gradient explosion, we truncate the descending curves and keep them unchanged for better visualization.}
\label{fig:allcomp}
%\vspace{-1mm}
\end{figure}

$\textbf{False-matching loss vs. cross entropy loss.}$ The two main loss functions to be compared with ours are permutation loss~\cite{wang2019learning} and permutation loss with Hungarian attention~\cite{yu2020learning}. We report accuracy/loss with training epoch in Fig.~\ref{fig:allcomp}. In experiments, the model with cross-entropy-type loss functions (Loss-P and Loss-H) always encounter the bad cases leading to gradient explosion in training stage, while our false matching loss (Loss-F) can avoid this issue. Besides, false matching loss is shown to do better against overfitting than the compared two loss functions.

\begin{table}[t!]
\scriptsize
\centering
\begin{tabular}{ccccc}
\toprule
\multicolumn{1}{c}{\begin{tabular}[c]{@{}c@{}}raw \\attributes\end{tabular}} & \multicolumn{1}{c}{\begin{tabular}[c]{@{}c@{}}unary \\geomteric prior\end{tabular}} & \multicolumn{1}{c}{\begin{tabular}[c]{@{}c@{}}pairwise \\structure context\end{tabular}} & \multicolumn{1}{c}{\begin{tabular}[c]{@{}c@{}}QC \\ optimization\end{tabular}} & accuracy \\
\midrule
$\surd$ & $\surd$ & $\surd$ & $\surd$ & 69.3\\
$\surd$ & $\surd$ & $\surd$ &  & 68.5\\
$\surd$ & $\surd$ &  &  & 68.3\\
$\surd$ &  &  &  &67.8\\
\bottomrule
\end{tabular}
\vspace{1mm}
\caption{Ablation study of qc-DG${\rm M}_1$ on Pascal VOC. The component been adopted is marked by a tick. ``QC optimization" is quadratic constrained-optimization.}
\label{table:ablation}
\vspace{-3mm}
\end{table}

\section{Conclusion}
In this work, we explicitly introduce quadratic constraint of graph structure into deep graph matching. To this end, unary geometric prior and pairwise structural context are considered for objective construction and a differentiable
optimization scheme is provided to approach the problem. Moreover, we focus on class imbalance that naturally exists in deep graph matching to propose our false matching loss. Experimental results show the competitive performance of our method. In future work, we plan to seek a more general form of quadratic constraint to the learning-based optimization for better matching.

%\newpage
{\small
\bibliographystyle{ieee_fullname}
\bibliography{egbib}

\begin{thebibliography}{10}\itemsep=-1pt

\bibitem{adams2011ranking}
Ryan~Prescott Adams and Richard~S Zemel.
\newblock Ranking via sinkhorn propagation.
\newblock {\em arXiv preprint arXiv:1106.1925}, 2011.

\bibitem{belongie2002shape}
Serge Belongie, Jitendra Malik, and Jan Puzicha.
\newblock Shape matching and object recognition using shape contexts.
\newblock {\em IEEE Transactions on Pattern Analysis and Machine Intelligence},
  24(4):509--522, 2002.

\bibitem{berg2005shape}
Alexander~C Berg, Tamara~L Berg, and Jitendra Malik.
\newblock Shape matching and object recognition using low distortion
  correspondences.
\newblock In {\em Proceedings of the IEEE Conference on Computer Vision and
  Pattern Recognition}, volume~1, pages 26--33. IEEE, 2005.

\bibitem{bottou2010large}
L{\'e}on Bottou.
\newblock Large-scale machine learning with stochastic gradient descent.
\newblock In {\em Proceedings of COMPSTAT'2010}, pages 177--186. Springer,
  2010.

\bibitem{bourdev2009poselets}
Lubomir Bourdev and Jitendra Malik.
\newblock Poselets: Body part detectors trained using 3d human pose
  annotations.
\newblock In {\em Proceedings of the IEEE International Conference on Computer
  Vision}, pages 1365--1372. IEEE, 2009.

\bibitem{brendel2011learning}
William Brendel and Sinisa Todorovic.
\newblock Learning spatiotemporal graphs of human activities.
\newblock {\em IEEE Transactions on Pattern Analysis and Machine Intelligence},
  38(9):1774--1789, 2016.

\bibitem{cho2013learning}
Minsu Cho, Karteek Alahari, and Jean Ponce.
\newblock Learning graphs to match.
\newblock In {\em Proceedings of the IEEE International Conference on Computer
  Vision}, pages 25--32, 2013.

\bibitem{cho2014finding}
Minsu Cho, Jian Sun, Olivier Duchenne, and Jean Ponce.
\newblock Finding matches in a haystack: A max-pooling strategy for graph
  matching in the presence of outliers.
\newblock In {\em Proceedings of the IEEE Conference on Computer Vision and
  Pattern Recognition}, pages 2083--2090, 2014.

\bibitem{conte2004thirty}
Donatello Conte, Pasquale Foggia, Carlo Sansone, and Mario Vento.
\newblock Thirty years of graph matching in pattern recognition.
\newblock {\em International Journal of Pattern Recognition and Artificial
  Intelligence}, 18(03):265--298, 2004.

\bibitem{egozi2012probabilistic}
Amir Egozi, Yosi Keller, and Hugo Guterman.
\newblock A probabilistic approach to spectral graph matching.
\newblock {\em IEEE Transactions on Pattern Analysis and Machine Intelligence},
  35(1):18--27, 2012.

\bibitem{everingham2010pascal}
Mark Everingham, Luc Van~Gool, Christopher~KI Williams, John Winn, and Andrew
  Zisserman.
\newblock The pascal visual object classes (voc) challenge.
\newblock {\em International Journal of Computer Vision}, 88(2):303--338, 2010.

\bibitem{fey2018splinecnn}
Matthias Fey, Jan Eric~Lenssen, Frank Weichert, and Heinrich M{\"u}ller.
\newblock Splinecnn: Fast geometric deep learning with continuous b-spline
  kernels.
\newblock In {\em Proceedings of the IEEE Conference on Computer Vision and
  Pattern Recognition}, pages 869--877, 2018.

\bibitem{fey2020deep}
Matthias Fey, Jan~E Lenssen, Christopher Morris, Jonathan Masci, and Nils~M
  Kriege.
\newblock Deep graph matching consensus.
\newblock {\em arXiv preprint arXiv:2001.09621}, 2020.

\bibitem{gasse2019exact}
Maxime Gasse, Didier Ch{\'e}telat, Nicola Ferroni, Laurent Charlin, and Andrea
  Lodi.
\newblock Exact combinatorial optimization with graph convolutional neural
  networks.
\newblock In {\em Proceedings of the Advances in Neural Information Processing
  Systems}, pages 15580--15592, 2019.

\bibitem{gold1996graduated}
Steven Gold and Anand Rangarajan.
\newblock A graduated assignment algorithm for graph matching.
\newblock {\em IEEE Transactions on Pattern Analysis and Machine Intelligence},
  18(4):377--388, 1996.

\bibitem{hartmanis1982computers}
Juris Hartmanis.
\newblock Computers and intractability: a guide to the theory of
  np-completeness.
\newblock {\em Siam Review}, 24(1):90, 1982.

\bibitem{ionescu2015training}
Catalin Ionescu, Orestis Vantzos, and Cristian Sminchisescu.
\newblock Training deep networks with structured layers by matrix
  backpropagation.
\newblock {\em arXiv preprint arXiv:1509.07838}, 2015.

\bibitem{jaggi2013revisiting}
Martin Jaggi.
\newblock Revisiting frank-wolfe: Projection-free sparse convex optimization.
\newblock In {\em Proceedings of the International Conference on Machine
  Learning}, number CONF, pages 427--435, 2013.

\bibitem{jiang2019glmnet}
Bo Jiang, Pengfei Sun, Jin Tang, and Bin Luo.
\newblock Glmnet: Graph learning-matching networks for feature matching.
\newblock {\em arXiv preprint arXiv:1911.07681}, 2019.

\bibitem{jiang2010linear}
Hao Jiang, X~Yu Stella, and David~R Martin.
\newblock Linear scale and rotation invariant matching.
\newblock {\em IEEE Transactions on Pattern Analysis and Machine Intelligence},
  33(7):1339--1355, 2010.

\bibitem{kipf2016semi}
Thomas~N Kipf and Max Welling.
\newblock Semi-supervised classification with graph convolutional networks.
\newblock {\em arXiv preprint arXiv:1609.02907}, 2016.

\bibitem{kuhn1955hungarian}
Harold~W Kuhn.
\newblock The hungarian method for the assignment problem.
\newblock {\em Naval Research Logistics Quarterly}, 2(1-2):83--97, 1955.

\bibitem{lacoste2015global}
Simon Lacoste-Julien and Martin Jaggi.
\newblock On the global linear convergence of frank-wolfe optimization
  variants.
\newblock In {\em Proceedings of the Advances in Neural Information Processing
  Systems}.

\bibitem{lawler1963quadratic}
Eugene~L Lawler.
\newblock The quadratic assignment problem.
\newblock {\em Management science}, 9(4):586--599, 1963.

\bibitem{leordeanu2005spectral}
Marius Leordeanu and Martial Hebert.
\newblock A spectral technique for correspondence problems using pairwise
  constraints.
\newblock In {\em Proceedings of the IEEE International Conference on Computer
  Vision}, volume~2, pages 1482--1489. IEEE, 2005.

\bibitem{leordeanu2009integer}
Marius Leordeanu, Martial Hebert, and Rahul Sukthankar.
\newblock An integer projected fixed point method for graph matching and map
  inference.
\newblock In {\em Advances in neural information processing systems}, pages
  1114--1122. Citeseer, 2009.

\bibitem{lin2017focal}
Tsung-Yi Lin, Priya Goyal, Ross Girshick, Kaiming He, and Piotr Doll{\'a}r.
\newblock Focal loss for dense object detection.
\newblock In {\em Proceedings of the IEEE International Conference on Computer
  Vision}, pages 2980--2988, 2017.

\bibitem{liu2014graph}
Zhi-Yong Liu, Hong Qiao, Xu Yang, and Steven~CH Hoi.
\newblock Graph matching by simplified convex-concave relaxation procedure.
\newblock {\em International Journal of Computer Vision}, 109(3):169--186,
  2014.

\bibitem{loiola2007survey}
Eliane~Maria Loiola, Nair Maria~Maia de Abreu, Paulo~Oswaldo Boaventura-Netto,
  Peter Hahn, and Tania Querido.
\newblock A survey for the quadratic assignment problem.
\newblock {\em European Journal of Operational Research}, 176(2):657--690,
  2007.

\bibitem{lowe2004distinctive}
David~G Lowe.
\newblock Distinctive image features from scale-invariant keypoints.
\newblock {\em International Journal of Computer Vision}, 60(2):91--110, 2004.

\bibitem{nair2010rectified}
Vinod Nair and Geoffrey~E Hinton.
\newblock Rectified linear units improve restricted boltzmann machines.
\newblock In {\em Proceedings of the International Conference on Machine
  Learning}, 2010.

\bibitem{rolinek2020deep}
Michal Rol{\'\i}nek, Paul Swoboda, Dominik Zietlow, Anselm Paulus, V{\'\i}t
  Musil, and Georg Martius.
\newblock Deep graph matching via blackbox differentiation of combinatorial
  solvers.
\newblock {\em arXiv preprint arXiv:2003.11657}, 2020.

\bibitem{russakovsky2015imagenet}
Olga Russakovsky, Jia Deng, Hao Su, Jonathan Krause, Sanjeev Satheesh, Sean Ma,
  Zhiheng Huang, Andrej Karpathy, Aditya Khosla, Michael Bernstein, et~al.
\newblock Imagenet large scale visual recognition challenge.
\newblock {\em International Journal of Computer Vision}, 115(3):211--252,
  2015.

\bibitem{schellewald2001evaluation}
Christian Schellewald, Stefan Roth, and Christoph Schn{\"o}rr.
\newblock Evaluation of convex optimization techniques for the weighted
  graph-matching problem in computer vision.
\newblock In {\em Joint Pattern Recognition Symposium}, pages 361--368.
  Springer, 2001.

\bibitem{simonyan2014very}
Karen Simonyan and Andrew Zisserman.
\newblock Very deep convolutional networks for large-scale image recognition.
\newblock {\em arXiv preprint arXiv:1409.1556}, 2014.

\bibitem{sinkhorn1967concerning}
Richard Sinkhorn and Paul Knopp.
\newblock Concerning nonnegative matrices and doubly stochastic matrices.
\newblock {\em Pacific Journal of Mathematics}, 21(2):343--348, 1967.

\bibitem{szeliski2010computer}
Richard Szeliski.
\newblock {\em Computer vision: algorithms and applications}.
\newblock Springer Science \& Business Media, 2010.

\bibitem{wfd2020frgm}
F.~D. {Wang}, N. {Xue}, Y. {Zhang}, G.~S. {Xia}, and M. {Pelillo}.
\newblock A functional representation for graph matching.
\newblock {\em IEEE Transactions on Pattern Analysis and Machine Intelligence},
  42(11):2737--2754, 2020.

\bibitem{wang2019learning}
Runzhong Wang, Junchi Yan, and Xiaokang Yang.
\newblock Learning combinatorial embedding networks for deep graph matching.
\newblock In {\em Proceedings of the IEEE International Conference on Computer
  Vision}, pages 3056--3065, 2019.

\bibitem{wang2019neural}
Runzhong Wang, Junchi Yan, and Xiaokang Yang.
\newblock Neural graph matching network: Learning lawler's quadratic assignment
  problem with extension to hypergraph and multiple-graph matching.
\newblock {\em arXiv preprint arXiv:1911.11308}, 2019.

\bibitem{wang2020combinatorial}
Runzhong Wang, Junchi Yan, and Xiaokang Yang.
\newblock Combinatorial learning of robust deep graph matching: an embedding
  based approach.
\newblock {\em IEEE Transactions on Pattern Analysis and Machine Intelligence},
  2020.

\bibitem{wang2020learning}
Tao Wang, He Liu, Yidong Li, Yi Jin, Xiaohui Hou, and Haibin Ling.
\newblock Learning combinatorial solver for graph matching.
\newblock In {\em Proceedings of the IEEE conference on computer vision and
  pattern recognition}, pages 7568--7577, 2020.

\bibitem{yan2016short}
Junchi Yan, Xu-Cheng Yin, Weiyao Lin, Cheng Deng, Hongyuan Zha, and Xiaokang
  Yang.
\newblock A short survey of recent advances in graph matching.
\newblock In {\em Proceedings of the 2016 ACM on International Conference on
  Multimedia Retrieval}, pages 167--174, 2016.

\bibitem{yan2015discrete}
Junchi Yan, Chao Zhang, Hongyuan Zha, Wei Liu, Xiaokang Yang, and Stephen~M
  Chu.
\newblock Discrete hyper-graph matching.
\newblock In {\em Proceedings of the IEEE Conference on Computer Vision and
  Pattern Recognition}, pages 1520--1528, 2015.

\bibitem{yu2020learning}
Tianshu Yu, Runzhong Wang, Junchi Yan, and Baoxin Li.
\newblock Learning deep graph matching with channel-independent embedding and
  hungarian attention.
\newblock In {\em Proceedings of the International Conference on Learning
  Representations}, volume~20, 2020.

\bibitem{zanfir2018deep}
Andrei Zanfir and Cristian Sminchisescu.
\newblock Deep learning of graph matching.
\newblock In {\em Proceedings of the IEEE Conference on Computer Vision and
  Pattern recognition}, pages 2684--2693, 2018.

\bibitem{zhou2015factorized}
Feng Zhou and Fernando De~la Torre.
\newblock Factorized graph matching.
\newblock {\em IEEE Transactions on Pattern Analysis and Machine Intelligence},
  38(9):1774--1789, 2015.

\bibitem{zhou2018graph}
Jie Zhou, Ganqu Cui, Zhengyan Zhang, Cheng Yang, Zhiyuan Liu, Lifeng Wang,
  Changcheng Li, and Maosong Sun.
\newblock Graph neural networks: A review of methods and applications.
\newblock {\em arXiv preprint arXiv:1812.08434}, 2018.

\end{thebibliography}
}

\end{document}